\documentclass[runningheads]{llncs}

\usepackage{eccv}

\usepackage{eccvabbrv}

\usepackage{graphicx}
\usepackage{booktabs}
\usepackage{xcolor}         % colors
\usepackage{enumitem}
\usepackage{multirow}
\usepackage{caption}           
\usepackage[table]{xcolor}
\usepackage{booktabs}
\usepackage{pifont}
\usepackage[utf8]{inputenc} % allow utf-8 input
\usepackage[T1]{fontenc}    % use 8-bit T1 fonts
\usepackage{url}            % simple URL typesetting
\usepackage{amsfonts}       % blackboard math symbols
\usepackage{nicefrac}       % compact symbols for 1/2, etc.
\usepackage{microtype}      % microtypography

\usepackage{wrapfig}
\usepackage{amsmath}
\usepackage{sidecap}
\usepackage{subcaption}

\usepackage{url}
\usepackage{tocloft}
\usepackage{titletoc}

\usepackage{balance}
\usepackage{enumitem}
\setlist[itemize]{label=\textbullet}

\usepackage[accsupp]{axessibility}  % Improves PDF readability for those with disabilities.

\usepackage{hyperref}

\usepackage{orcidlink}

\begin{document}

% ---------------------------------------------------------------
% TODO REVIEW: Replace with your title
\title{Stay Unique, Stay Efficient: Preserving Model Personality in Multi-Task Merging} 

% TODO REVIEW: If the paper title is too long for the running head, you can set
% an abbreviated paper title here. If not, comment out.
\titlerunning{Preserving Model Personality in Multi-Task Merging}

% TODO FINAL: Replace with your author list. 
% Include the authors' OCRID for the camera-ready version, if at all possible.
\author{Kuangpu Guo\inst{1,2}\orcidlink{0009-0002-7267-0428} \and
Aijing Yu\inst{4}\orcidlink{0000-0002-4782-9858} \and
Jian Liang\inst{2,3}\thanks{Corresponding author.}\orcidlink{0000−0003−3890−1894} \and 
Yuhe Ding\inst{5}\orcidlink{0000-0002-7572-3049} \and\\
Zilei Wang\inst{1}\orcidlink{0000-0003-1822-3731} \and
Ran He\inst{2,3}\orcidlink{0000−0002−3807−991X} \and
Tieniu Tan\inst{2,3,6}\orcidlink{0000-0002-1808-2169}}

% TODO FINAL: Replace with an abbreviated list of authors.
\authorrunning{Guo. et al.}
% First names are abbreviated in the running head.
% If there are more than two authors, 'et al.' is used.

% TODO FINAL: Replace with your institution list.
\institute{University of Science and Technology of China \and
NLPR \& MAIS, Institute of Automation, Chinese Academy of Sciences \and
School of Artificial Intelligence, University of Chinese Academy of Sciences \and 
% School of Cyber Security, University of Chinese Academy of Sciences \and
Institute of Information Engineering, Chinese Academy of Sciences \and
Anhui University \and 
Nanjing University \\
\email{gkp@mail.ustc.edu.cn, liangjian92@gmail.com}
}

\maketitle

\begin{abstract}
Model merging has emerged as a promising paradigm for enabling multi-task capabilities without additional training. 
However, traditional basic merging methods often experience performance degradation due to parameter conflicts, even when applied to similar tasks. 
While recent personalized merging frameworks successfully preserve task-specific information to maintain performance, they typically incur storage overhead.
In this paper, we propose \textbf{D}ecomposition, \textbf{T}hresholding, and \textbf{S}caling (\textbf{DTS}), an approximation-based personalized merging framework that pushes task-specific storage efficiency.
DTS first applies singular value decomposition to the task-specific information and retains only a small subset of singular values and vectors.
It then introduces a novel thresholding strategy that partitions singular vector elements into groups and assigns a scaling factor to each group.
To enable generalization to unseen tasks, we further extend DTS with a variant that fuses task-specific information in a data-free manner based on the semantic similarity of task characteristics.
Extensive experiments demonstrate that DTS consistently outperforms state-of-the-art baselines while requiring only 1\% extra storage per task.
Furthermore, experiments on unseen tasks show that the DTS variant achieves significantly better generalization performance.
Our code is available at \url{https://github.com/krumpguo/DTS}.
\end{abstract}

\section{Introduction}
\label{sec:intro}
Adapting pre-trained models to diverse tasks via fine-tuning~\cite{wolf2019huggingface, ye2023merging, li2025model, yang2025mix} often leads to prohibitive deployment costs. 
Unlike data-intensive multi-task learning~\cite{caruana1997multitask, zhang2021survey}, model merging~\cite{zhaomerging, limap, zhou2024hm3, hyun2025multi, qiu2025mingle} offers a data-free and optimization-free alternative by directly fusing parameters from multiple task-specific models into a unified representation.

\begin{figure}[t]
  \centering
  \hspace*{\fill}
  \setlength{\abovecaptionskip}{0.cm}
  \begin{subfigure}{0.32\textwidth}
    \centering
    \includegraphics[width=\linewidth]{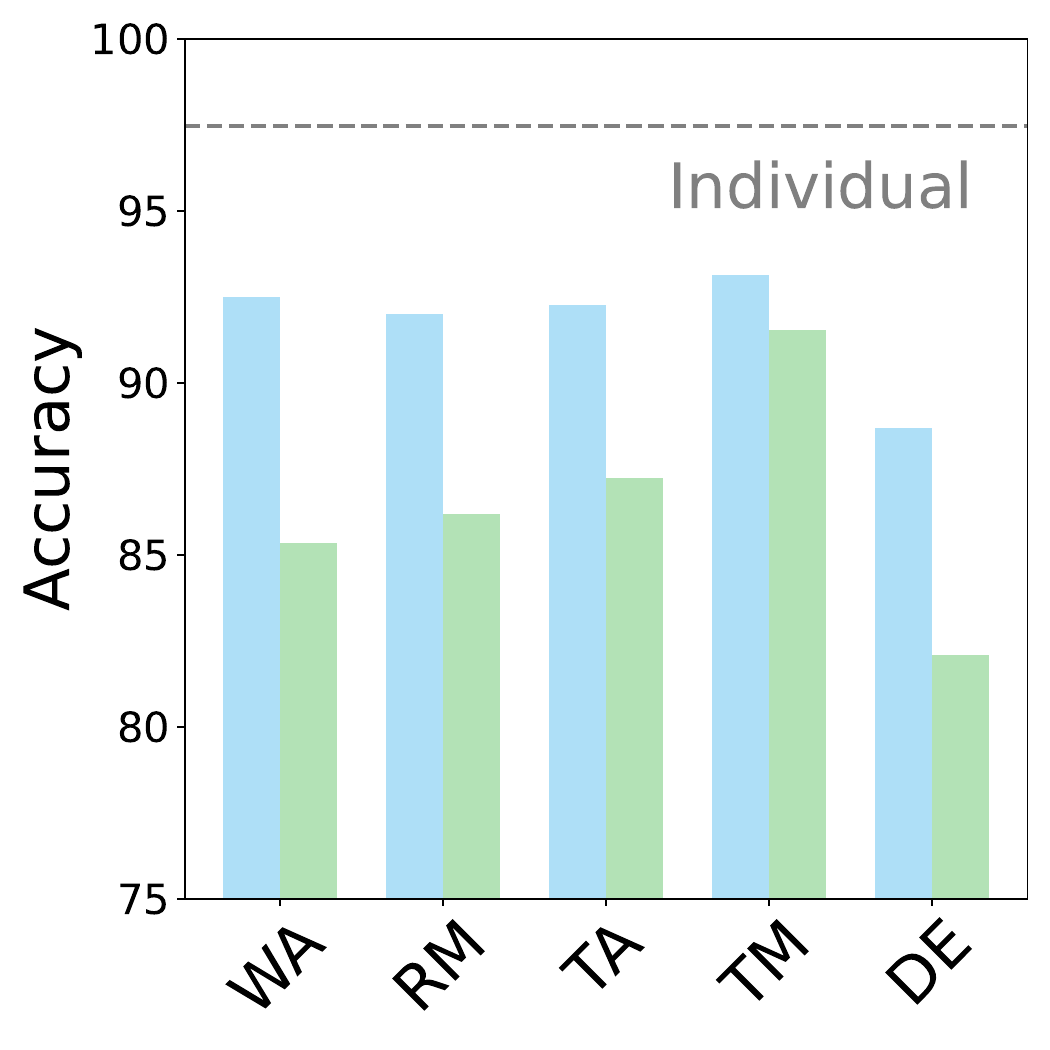}
    \caption{SVHN}
    \label{fig:plot_SVHN}
  \end{subfigure}
  \hfill
  \begin{subfigure}{0.32\textwidth}
    \centering
    \includegraphics[width=\linewidth]{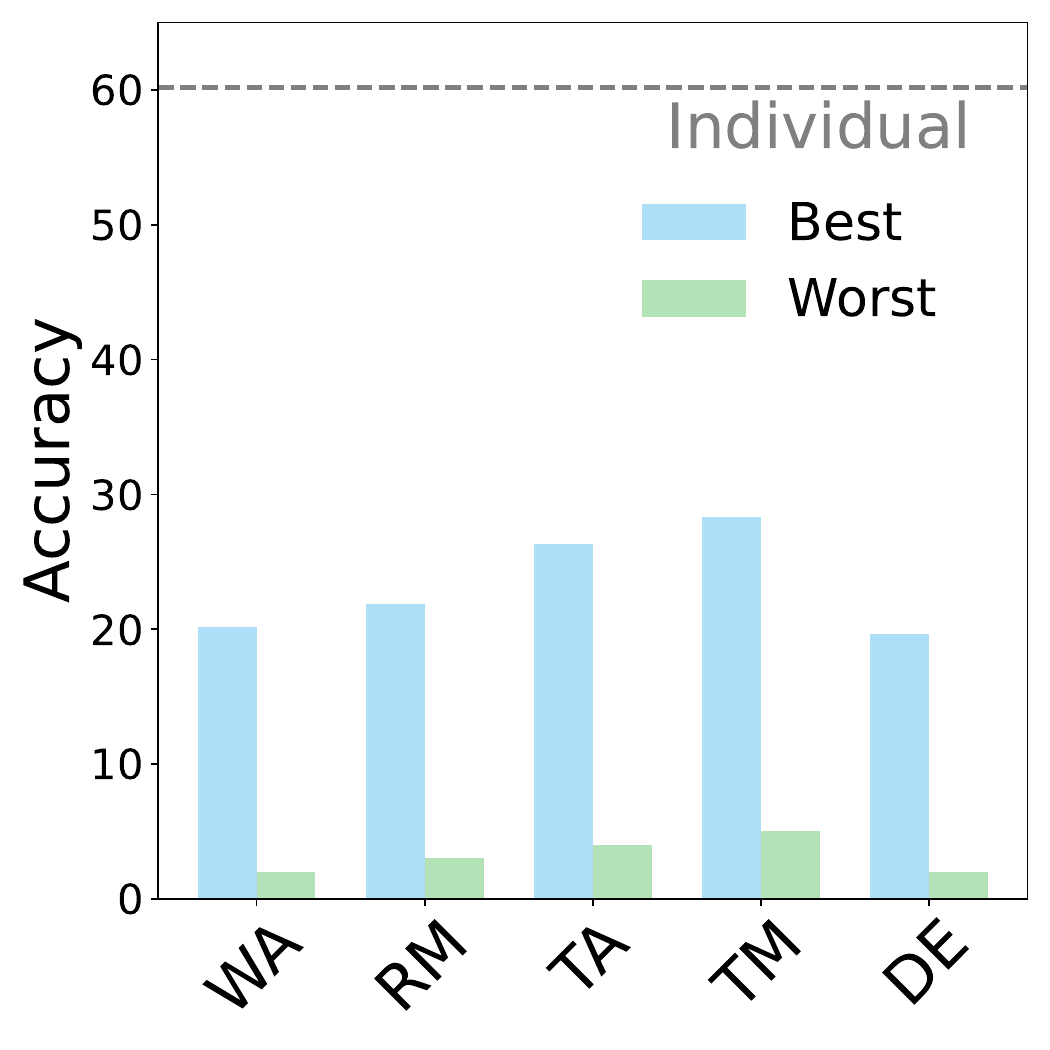}
    \caption{CoLA}
    \label{fig:toy_CoLA}
  \end{subfigure}
  \hspace*{\fill}
  \caption{Performance of the pairwise merged model: we pairwise merge models fine-tuned on SVHN~\cite{netzer2011reading} or CoLA~\cite{warstadt2019neural} with those fine-tuned on the other seven datasets, reporting the best and worst performance across these combinations.WA, RM, TA, TM, and DE refer to Weight-Averaging~\cite{wortsman2022model}, RegMean~\cite{jin2022dataless}, Task-Arithmetic~\cite{ilharco2022editing}, Ties-Merging~\cite{yadav2023ties}, and DARE~\cite{yu2024language}, respectively.}
  \label{fig:same_task_1}
\end{figure}

Traditional basic model merging methods~\cite{ilharco2022editing}  strive to compress all capabilities into a single, unified parameter body. 
However, these methods inherently suffer from severe parameter conflicts~\cite{yadav2023ties, qi2024less}. 
To investigate the depth of this issue, we examine the performance degradation from the perspective of task similarity. 
Specifically, we conduct pairwise model merging by combining the model fine-tuned on the SVHN dataset with models fine-tuned on each of the other seven datasets in the visual benchmark.
Fig.~\ref{fig:same_task_1} (a) shows the best and worst performance of the merged models when evaluated on SVHN, corresponding to merging with the most similar task (MNIST) and the most dissimilar task (EuroSAT), respectively.
Surprisingly, we observe significant performance drops even when merging models trained on the most similar tasks.
These results highlight the persistent nature of cross-task parameter conflicts and underscore the fundamental limitation of enforcing a single shared parameter set across all tasks.

To address this issue, a second paradigm, personalized model merging (e.g., WEMOE~\cite{tang2024merging}, EMR-Merging~\cite{huang2024emr}), has recently gained traction. 
By recognizing the necessity of preserving task-specific information, these approaches retain a subset of task-specific parameters or employ dynamic routing modules conditioned on task IDs. 
While they successfully mitigate interference and achieve performance strictly bounded by individual fine-tuning, existing personalized merging frameworks often incur prohibitive storage overhead.
Storing dense expert components or complex routing networks negates the primary advantage of parameter efficiency in model merging, severely limiting their deployment in resource-constrained environments.

To bridge this gap, we propose \textbf{DTS}, an approximation-based personalized model merging method that leverages \textbf{D}ecomposition, \textbf{T}hresholding, and \textbf{S}caling to efficiently preserve task-specific information while minimizing storage overhead.
DTS applies singular value decomposition (SVD)~\cite{lange2010singular} to each parameter matrix in the task-specific information, retaining only a small subset of the filtered singular values and vectors.
This selective retention significantly reduces storage requirements while maintaining essential task information.
To further reduce storage costs, we introduce a novel thresholding strategy to approximate the singular vectors.
Unlike prior works~\cite{qi2024less, kim2025task}, which either retain a subset of elements using a binary mask or directly quantize the parameters, we argue that such coarse processing may discard important fine-grained information.
Instead, we propose a novel thresholding strategy that thresholds the elements of each singular vector into four groups and assigns a scaling factor for each group, as illustrated in Fig.~\ref{fig:compare1}.
This strategy improves the approximation for the original task-specific information.
In addition to applying DTS to task vectors (DTS-T), we introduce a new form of task-specific information termed the difference vector, which captures the parameter difference between individually fine-tuned and merged models.
We apply DTS to this difference vector, referring to it as DTS-D.

To enable generalization to unseen tasks, we propose a variant of DTS that adaptively integrates task-specific information from seen tasks.
Unlike prior methods that require additional data or trainable routers to determine merging weights, our approach computes weights based on the average semantic similarity between seen and unseen task characteristics.
This design is entirely data-free, offering both efficiency and scalability.
Experimental results demonstrate that DTS consistently outperforms state-of-the-art baselines on standard multi-task model merging benchmarks, achieving strong performance with only 1\% extra storage per task.
Furthermore, on unseen tasks, the DTS variant delivers superior generalization performance compared to all competing methods.
Our contributions can be summarized as follows:

\begin{itemize}[leftmargin=15pt, itemsep=0pt, topsep=0pt]
\item We propose \textbf{DTS}, a lightweight approximation-based personalized merging method that addresses performance degradation from parameter conflicts by preserving task-specific information.
\item We introduce DTS-T and DTS-D variants, 
which apply DTS on task vectors and difference vectors, respectively. 
Additionally, we present a variant of DTS that fuses task-specific information based on the semantic similarity of task characteristics, enabling effective generalization to unseen tasks.
\item  DTS consistently outperforms state-of-the-art baselines on standard benchmarks with only $\approx$1\% extra storage per task.
Furthermore, experiments on unseen tasks show that the DTS variant achieves superior generalization performance compared to all baseline methods, highlighting its effectiveness in both seen and unseen task settings.
\end{itemize}

\section{Related Work}
\label{related_work}
Model merging~\cite{ainsworth2022git, ilharco2022editing, touvron2023llama, zhang2025beyond, chen2025fw, ding2025harmonizing}, which aims to fuse multiple fine-tuned models into a single comprehensive model, has attracted increasing attention with the release of numerous publicly available model checkpoints~\cite{wolf2019huggingface, dodge2020fine, jordan2023repair}.
Model merging significantly reduces storage and deployment cost by unifying multiple models into a single one, without additional training~\cite{ yu2024language, he2024gradient, liu20251bit, suncat}.
Depending on whether the merged model is static or tailored per task, existing methods can be categorized into basic and personalized model merging methods.

\subsection{Basic Model Merging}
Basic model merging~\cite{ilharco2022editing, wortsman2022model, jin2022dataless, matena2022merging, yadav2023ties, yu2024language, davari2024model, gargiulo2025task, marczakno} focuses on universal strategies for merging fine-tuned models into a single unified model.
A canonical example is Weight-Averaging~\cite{wortsman2022model}, which simply averages parameters from different tasks.
Task-Arithmetic~\cite{ilharco2022editing} introduces task vectors that yield a better merged model.
Building on this idea, DARE~\cite{yu2024language} and Ties-Merging~\cite{yadav2023ties} propose pruning and scaling task vectors, assuming that not all parameters contribute equally.
AdaMerging~\cite{yang2023adamerging} introduces adaptive learning to automate coefficient selection but incurs additional training cost.
Other approaches like Fisher-Merging~\cite{jin2022dataless} and RegMean~\cite{matena2022merging} compute merging coefficients via Fisher information or inner product matrices.
However, basic model merging often results in parameter conflicts and a loss of task-specific information, leading to a substantial performance gap compared to individual models~\cite{leemitigating,stoicamodel, yangcontinual, zhong2025aim}.
Therefore, we explore the personalized model merging in this work.

\subsection{Personalized Model Merging}
Personalized model merging~\cite{muqeeth2023soft, wang2024localizing, lu2024twin, tang2024merging, huang2024emr, zhu2025remedy} enhances the merged model with task-specific components to boost performance across a variety of tasks.
SMEAR~\cite{muqeeth2023soft} and Twin-Merging~\cite{lu2024twin} store full model parameters for each task and employ routing mechanisms to perform weighted parameter fusion based on expert distributions. 
WEMOE~\cite{tang2024merging} introduces test-time adaptation by merging most weights while converting MLP layers into a mixture-of-experts (MoE) module.
DaWin~\cite{oh2024dawin} similarly retains full models and uses entropy over unlabeled test samples to assess task relevance. 
While effective, these methods typically require access to training data or incur high storage costs. 
In contrast, EMR-Merging~\cite{huang2024emr} selects a unified base model and generates lightweight, data-free modulators for each task. 
T-Switch~\cite{qi2024less} further reduces memory by storing task vectors in a binarized form. 
FREE-Merging~\cite{zheng2025free} proposes a lightweight task-specific expert module that dynamically compensates for information loss during merging.
Nonetheless, existing personalized methods often suffer from either data dependency or excessive storage demands.
In this work, we aim to retain essential task-specific information with minimal memory overhead, mitigating parameter conflicts without relying on additional data or training.

\section{Method}
\label{method}

\begin{figure*}[t]                       
  \centering                           
  \includegraphics[width=\linewidth]{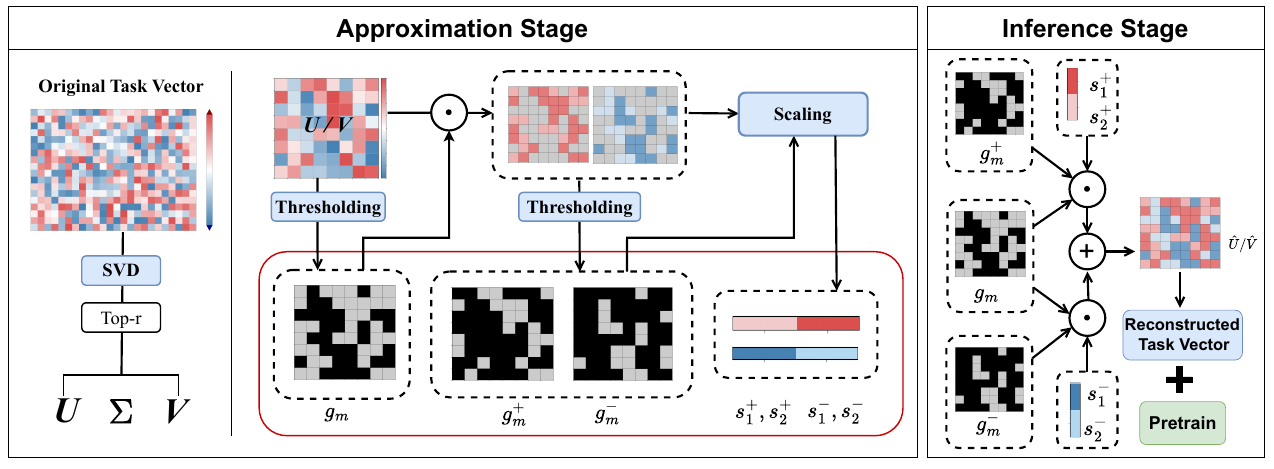}
  \caption{Framework overview. In the approximation stage, we first apply singular value decomposition to the task-specific information. Next, a novel thresholding strategy is used to group the elements of each singular vector, followed by computing a scaling factor for each group. During inference, the task-specific information is reconstructed using the approximated singular vectors for each task.   
           }
  \label{fig:pipeline}               
\end{figure*}

\subsection{Problem Formulation}
In this paper, following the setup of prior model merging works~\cite{ilharco2022editing, yadav2023ties, yu2024language, qi2024less}, we consider a scenario involving $N$ tasks with corresponding datasets $\{\mathcal{D}_n\}_{n=1}^{N}$, where each sample $(\boldsymbol{x}_n, y_n) \in \mathcal{D}_n$ belongs to the $n$-th task. 
Let $f_{\boldsymbol{\theta}_0}$ denote a pre-trained model with parameters $\boldsymbol{\theta}_0 \in \boldsymbol{\Theta}$, and let $f_{\boldsymbol{\theta}_1}, \ldots, f_{\boldsymbol{\theta}_N}$ represent task-specific models fine-tuned on each $\mathcal{D}_n$, where $\boldsymbol{\theta}_n$ represents the fine-tuned weights for the $n$-th task.
Following Task-Arithmetic~\cite{ilharco2022editing}, the task vector $\boldsymbol{\tau}_n$ for the $n$-th task is defined as $\boldsymbol{{\tau}}_n = \boldsymbol{{\theta}}_n - \boldsymbol{{\theta}}_0$, which serves as a widely adopted representation for capturing task-specific parameter information.

While the standard task vector $\boldsymbol{\tau}_n$ effectively isolates task-specific parameter shifts, it strictly relies on the availability of the original pre-trained weights $\boldsymbol{\theta}_0$. 
In many practical open-source scenarios, practitioners might only have access to a collection of fine-tuned models or a coarsely merged baseline $\boldsymbol{\theta}_m$. 
To address this constraint, we introduce a new form of task-specific information termed the \textbf{difference vector}, defined as $\boldsymbol{d}_n = \boldsymbol{\theta}_n - \boldsymbol{\theta}_m$, where $\boldsymbol{\theta}_m$ is the parameter set obtained via a basic merging method such as Ties-Merging~\cite{yadav2023ties}.
Beyond serving as a practical alternative when $\boldsymbol{\theta}_0$ is inaccessible, $\boldsymbol{d}_n$ captures the personalized residual that deviates from the shared common knowledge embedded in $\boldsymbol{\theta}_m$. 
By explicitly modeling these residuals relative to a task-agnostic baseline, this formulation provides a robust anchor for generalizing to unseen tasks.

As an illustrative example, we consider task vectors as task-specific information. The goal of model merging is to combine the set $\{\boldsymbol{\tau}_n\}_{n=1}^{N}$ with the pre-trained model to produce a merged model that performs well on the union of all task datasets, $\mathcal{D} = \bigcup_{n=1}^{N} \mathcal{D}_n$, formulated as:
\begin{equation}
\begin{aligned}
\text{min}  \ \mathbb{E}_{(\boldsymbol{x}, y) \in \mathcal{D}} \ \mathcal{L} \left( f_{\boldsymbol{\theta}_{m}^*} (\boldsymbol{x}), y \right), \  \boldsymbol{\theta}_{m}^* = \text{Merge} (\boldsymbol{{\theta}}_0, \{ \boldsymbol{\tau}_n \}_{n=1}^{N}).
\end{aligned}
\end{equation}
Here, $\text{Merge}(\cdot)$ denotes a general merging function.
For instance, Task Arithmetic~\cite{ilharco2022editing} takes the form $\text{Merge}(\boldsymbol{\theta}_0, \{ \boldsymbol{\tau}_n \}_{n=1}^{N}) = \boldsymbol{\theta}_0 + \sum_{n=1}^{N} \gamma_n \boldsymbol{\tau}_n$.
More advanced personalized strategies, such as generating task-specific modulators~\cite{huang2024emr}, can also be employed to merge models across diverse tasks.
\textit{Following prior studies~\cite{ilharco2022editing, yu2024language, yang2023adamerging, lu2024twin, huang2024emr, wang2024localizing, qi2024less}, we also assume that the information of the test task (e.g., task ID) is available during evaluation.}

\subsection{Decomposition, Thresholding, and Scaling}
Previous studies~\cite{zhou2024metagpt,marczakno,yancalm, wu2025importance} have shown that the merged model often suffers from notable performance degradation compared to its fine-tuned counterparts, primarily due to parameter conflicts among the fine-tuned models.
In this work, we examine this issue through the lens of task similarity.
Specifically, we perform pairwise model merging by combining a model fine-tuned on SVHN~\cite{netzer2011reading} with models fine-tuned on each of the other seven datasets in the benchmark.

Fig.~\ref{fig:same_task_1} (a) shows the best and worst SVHN performance among the merged models, corresponding to merging with the most similar task (MNIST~\cite{deng2012mnist}) and the most dissimilar one, respectively.
Surprisingly, substantial performance drops are observed even when merging models from similar tasks.
A similar pattern emerges in natural language processing tasks, as illustrated in Fig.~\ref{fig:same_task_1} (b). 
These findings suggest that parameter conflicts are inherent, even across semantically related tasks, and underscore the importance of preserving personalized information to maintain model performance. 
This insight motivates our objective: to retain complete task-specific information while minimizing extra storage overhead.

\begin{figure}[t]
  \centering
  \hspace*{\fill}
  \setlength{\abovecaptionskip}{0.1cm}
  \begin{subfigure}{0.21\textwidth}
    \centering
    \includegraphics[width=\linewidth]{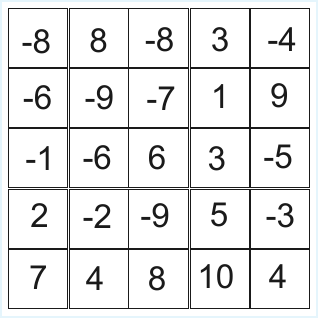}
    \caption{Singular vector}
    \label{fig:toy_orig}
  \end{subfigure}
  \hfill
  \begin{subfigure}{0.21\textwidth}
    \centering
    \includegraphics[width=\linewidth]{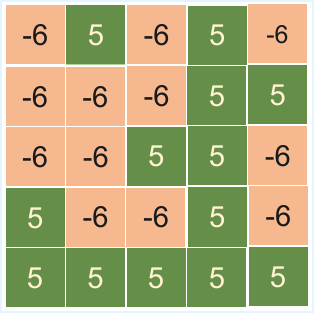}
    \caption{Binarization}
    \label{fig:toy_bin}
  \end{subfigure}
  \hfill
  \begin{subfigure}{0.21\textwidth}
    \centering
    \includegraphics[width=\linewidth]{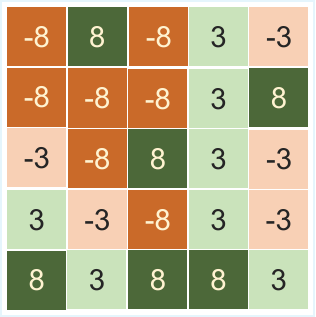}
    \caption{Thresholding}
    \label{fig:toy_thr}
  \end{subfigure}
  \hspace*{\fill}
  \caption{A toy example of our thresholding strategy. Thresholding provides a more fine-grained approximation by further partitioning positive and negative values into large/small groups.}
  \label{fig:compare1}
% \vspace*{-14pt}
\end{figure}

In this paper, we propose preserving the task-specific information of individual models. 
However, storing full model parameters incurs substantial memory overhead, making this approach impractical in resource-constrained settings. 
To address this issue, we introduce \textbf{D}ecomposition, \textbf{T}hresholding, and \textbf{S}caling (\textbf{DTS}), a personalized method that approximates task-specific information while preserving its effectiveness, as illustrated in Fig.~\ref{fig:pipeline}.
To establish a unified compression pipeline for both variants (DTS-T and DTS-D), let $\boldsymbol{W}_n \in \{\boldsymbol{\tau}_n, \boldsymbol{d}_n\}$ denote the target task-specific deviation matrix that requires storage. 
For layers whose parameters have more than two dimensions (e.g., convolutional kernels), we first reshape the tensor into a 2D matrix before processing.
Specifically, we first apply truncated singular value decomposition to the target matrix and retain only the top-$r$ singular values, as follows:
\begin{equation}
\boldsymbol{U}_n, \boldsymbol{\Sigma}_n, \boldsymbol{V}_n = \text{SVD}_r(\boldsymbol{W}_n),
\end{equation}
where $\boldsymbol{U}_n$ and $\boldsymbol{V}_n$ denote the left and right singular vector matrices, and $\boldsymbol{\Sigma}_n$ contains the retained singular values. The sparsity coefficient $r$ is the proportion of singular values preserved.

To further reduce storage, we threshold the decomposed components. 
Unlike previous work~\cite{qi2024less}, which uses a simple binary mask to retain a subset of elements as shown in Fig.~\ref{fig:compare1} (b), we argue that this coarse masking approach may result in the loss of fine-grained information. 
Instead, we propose a novel thresholding strategy that partitions the elements of each singular vector into four groups and computes a scaling factor for each group as shown in Fig.~\ref{fig:compare1} (c). 
Specifically, taking the left singular vector matrix $\boldsymbol{U}_n$ as an example, we first mark the sign of each element as follows:
\begin{equation}
\label{eq:positive_negative}
g(U_{n,j}) = 
\begin{cases}
1, & \text{if } U_{n,j} > 0, \\
0, & \text{otherwise},
\end{cases}
\end{equation}
where $U_{n,j}$ is the $j$-th element of $\boldsymbol{U}_n$. 
This thresholding function encodes the sign of each parameter using only 1 bit, dividing the parameters into positive and negative groups.
To preserve more fine-grained information, we further threshold the positive and negative values into two subgroups, respectively. 
Taking the positive values as an example:
\begin{equation}
g^{+}(U_{n,j}) = 
\begin{cases}
\text{1}, & \text{if } U_{n,j} > \lambda, \\
\text{0}, & \text{otherwise},
\end{cases}
\end{equation}
where $\lambda$ is the median of the positive entries in $\boldsymbol{U}_n$.
This extended thresholding partitions positive elements into "large" and "small" groups.
The same strategy is applied to the negative values.
During the inference stage, we reconstruct the positive portion of $\boldsymbol{U}_n$ as:
\begin{equation}
\begin{aligned}
\hat{\boldsymbol{U}}^{+}_n = s^+_1 \cdot g(\boldsymbol{U}_n) \odot g^{+}(\boldsymbol{U}_n)
+ s^+_2 \cdot g(\boldsymbol{U}_n) \odot (1 - g^{+}(\boldsymbol{U}_n)),
\end{aligned}
\end{equation}
where $\odot$ denotes element-wise multiplication. 
The scaling factors $s^+_1$ and $s^+_2$  align the magnitude of the approximated vectors with the original components.

The design of DTS is theoretically grounded in optimal low-rank approximation and highly efficient scalar quantization. 
According to the Eckart-Young-Mirsky theorem~\cite{eckart1936approximation}, the truncated SVD reconstruction provides the theoretically optimal low-rank approximation. 
The truncation error is strictly bounded by the discarded singular values, effectively filtering out unstructured parameter noise. 
Furthermore, our 4-group thresholding serves as a median-based piecewise constant approximation.
Given these fixed median splits, we aim to achieve the conditional minimum Mean Squared Quantization Error (MSQE) within each subgroup $\mathcal{S}_k$.
The optimal scalar $s_k$ that minimizes this $L_2$ residual is the arithmetic mean of the absolute values of the elements within $\mathcal{S}_k$.
For efficient engineering implementation using tensor operations, we compute this arithmetic mean using the $L_1$ norm:
\begin{equation}
\begin{aligned}
s^+_1 =  \frac{\| \boldsymbol{U}_n \odot g(\boldsymbol{U}_{n}) \odot g^{+}(\boldsymbol{U}_{n})\|_1}{\| g(\boldsymbol{U}_{n}) \odot g^{+}(\boldsymbol{U}_{n})\|_1}, \
s^+_2 =  \frac{\|\boldsymbol{U}_n \odot g(\boldsymbol{U}_{n}) \odot (1 - g^{+}(\boldsymbol{U}_n))\|_1}{\|g(\boldsymbol{U}_{n}) \odot (1 - g^{+}(\boldsymbol{U}_n))\|_1}.
\end{aligned}
\end{equation}
The numerator calculates the sum of absolute values within the subgroup, and the denominator calculates the exact cardinality. 
This formulation leverages the $L_1$ norm to compute the exact arithmetic mean, ensuring optimal local magnitude alignment under the median-based partition.
In Fig.~\ref{fig:compare1}, the computed values of $s^+_1$ and $s^+_2$ are 8 and 3, respectively. 
We apply the identical procedure to the negative values and to obtain the approximated $\hat{\boldsymbol{V}}_n$. 
For one-dimensional parameters (e.g., biases), we skip SVD and directly apply thresholding and scaling.

Crucially, this 4-group configuration allocates precisely 1 bit for the sign and 1 bit for the magnitude level per element, equating to a 2-bit representation (a 16$\times$ compression ratio). 
Finally, during the inference phase, the task-specific parameters are dynamically reconstructed by adding the approximated deviation back to their corresponding reference base model:
\begin{equation}
\label{reconstruction}
\hat{\boldsymbol{\theta}}_n = \boldsymbol{\theta}_{\text{ref}} + \hat{\boldsymbol{U}}_n \boldsymbol{\Sigma}_n \hat{\boldsymbol{V}}_n^T,
\end{equation}
where $\boldsymbol{\theta}_{\text{ref}} = \boldsymbol{\theta}_0$ for the DTS-T variant, and $\boldsymbol{\theta}_{\text{ref}} = \boldsymbol{\theta}_m$ for the DTS-D variant. 
As discussed, this process preserves essential task-specific information while mitigating parameter conflicts, requiring only $\approx 1\%$ extra storage per task.

\subsection{Extending DTS for Unseen Tasks}
In practical deployments, merged models frequently encounter unseen tasks. 
While the base merged model can serve as a generic initialization, it does not account for task similarities, which limits its generalization performance. 
The objective here is to adaptively transfer the preserved task-specific residuals (e.g., the approximated difference vectors $\hat{\boldsymbol{d}}_n$) to novel tasks without requiring additional training data.

To achieve this, we utilize the semantic meta-information of tasks as a training-free metric for parameter transferability. 
Unlike existing adaptation methods that rely on target-domain data to compute merging weights via gradient updates, our approach leverages semantic similarity to enable data-free generalization.

For clarity, we illustrate the approach using the DTS-D variant. 
Taking classification tasks as an example, where class names serve as task-specific characteristics, we encode them with a pretrained text encoder to generate class name embeddings. 
The task-level semantic representation is obtained by averaging the embeddings of all class names within a given task. 
For an unseen task, we compute the mean embedding $E_u$ over its class names. 
Given the embeddings $\{E_n\}_{n=1}^{N}$ of the seen tasks, the dynamic merging process is formulated as:
\begin{equation}
\begin{aligned}
\hat{\boldsymbol{\theta}}_u = \boldsymbol{\theta}_m + \sum_{n=1}^N \gamma_n \cdot \hat{\boldsymbol{d}}_n,  \ \gamma_n = \frac{\text{cos}(E_u, E_n)}{\sum_{k=1}^N \text{cos}(E_u, E_k)},
\end{aligned}
\end{equation}
where $\text{cos}(\cdot, \cdot)$ denotes the cosine similarity. 
The identical strategy applies to the DTS-T variant by substituting $\hat{\boldsymbol{d}}_n$ with $\hat{\boldsymbol{\tau}}_n$ and replacing $\boldsymbol{\theta}_m$ with $\boldsymbol{\theta}_0$. 
For generation tasks, task characteristics are obtained by encoding the corresponding task instructions or dataset descriptions with a language model.

Unlike existing adaptation mechanisms that require newly collected labeled data and iterative backpropagation, our proposed variant operates in a strictly zero-shot and data-free manner. 
The computational overhead introduced by the text encoder is minimal, as all semantic task embeddings are extracted offline prior to deployment. 
Consequently, this strategy provides a highly efficient adaptation mechanism for unseen tasks without introducing any additional training cost or online inference latency.

\begin{table*}[t]
  \centering
  \setlength{\tabcolsep}{3.5pt}
    \renewcommand{\arraystretch}{0.92}
  \caption{Multi-task performance (\%) when merging ViT-B/32 models on eight tasks. ADR refers to the Accuracy Drop Rate, and AMR denotes the Additional Memory Rate. The best result is \textbf{highlighted}, and the second-best result is \underline{underlined}.
  \textbf{Per} indicates whether the merging is a personalized method.}
  \label{table_vit-b-32}
  \resizebox{\textwidth}{!}{
  \begin{tabular}{lccccccccccccc}
    \toprule \rowcolor{gray!10}
    Method &  \textbf{Per} & SUN397 & Cars & RESISC45 & EuroSAT & SVHN & GTSRB & MNIST & DTD 
      & Avg. $\uparrow$ & ADR $\downarrow$ & AMR $\downarrow$\\
    \midrule	 	  	 	 	 	 
    \rowcolor{cyan!10}
    Individual     & -          & 74.49 & 77.73 & 98.22 & 99.80 & 97.46 & 98.73 & 99.69 & 79.36 & 90.69 & -- & --    \\ \hline
    \rowcolor{purple!10} 
    Weight-Averaging~\cite{wortsman2022model}     & \ding{55}    & 65.35 & 63.41 & 71.42 & 71.69 & 64.20 & 52.82 & 87.56 & 50.18 & 65.83  & 27.42 & 0.00    \\
    \rowcolor{purple!10}
    Fisher-Merging~\cite{matena2022merging}   & \ding{55}        & 68.69 & 69.21 & 70.73 & 66.41 & 72.91 & 51.17 & 87.94 & 59.99 & 68.38 & 24.61 & 0.00    \\
    \rowcolor{purple!10}
    RegMean~\cite{jin2022dataless}      & \ding{55}            & 65.35 & 63.53 & 75.61 & 78.66 & 78.10 & 67.49 & 93.75 & 52.02 & 71.81 & 20.71 & 0.00     \\
    \rowcolor{purple!10}
    Task-Arithmetic~\cite{ilharco2022editing}  & \ding{55}         & 54.78 & 54.98 & 67.69 & 78.70 & 80.21 & 69.68 & 97.34 & 50.37 & 69.22 & 23.68 & 0.00     \\
    \rowcolor{purple!10}
    Ties-Merging~\cite{yadav2023ties}    & \ding{55}         & 64.17 & 64.43 & 76.31 & 76.62 & 81.28 & 69.37 & 96.53 & 54.52 & 72.90 & 19.62 & 0.00     \\
    \rowcolor{purple!10}
    DARE~\cite{yu2024language}       & \ding{55}               & 64.76 & 63.08 & 71.02 & 70.70 & 62.04 & 50.68 & 86.17 & 50.64 & 64.89 & 28.45  & 0.00   \\
    \rowcolor{purple!10}
    AdaMerging~\cite{yang2023adamerging}     & \ding{55}           & 64.44 & 68.05 & 79.31 & 93.80 & 87.06 & 91.93 & 97.56 & 59.11 & 80.16 & 11.62 & 0.00     \\
    \rowcolor{purple!10}
    AdaMerging++~\cite{yang2023adamerging}      & \ding{55}        & 66.61 & 68.34 & 82.28 & 94.11 & 89.54 & 89.01 & 98.18 & 60.66 & 81.09 & 10.58  & 0.00    \\ \hline
    \rowcolor{blue!10}  
    SVD  & \ding{51} &  71.18 & 71.60 & 96.95 & 99.64 & 97.24 & 98.02 & \underline{99.66} & 77.58 & 88.98  &1.88 & 5.87 \\   
    \rowcolor{blue!10}
    Twin-Merging~\cite{lu2024twin}       & \ding{51}      & 71.56 & 68.78 & 89.97 & 72.11 & 96.65 & 93.35 & \underline{99.66} & 72.50 & 83.07 & 8.41  & 100.0 \\ 
    \rowcolor{blue!10} 
    WEMOE~\cite{tang2024merging}   & \ding{51}                 & 73.92 & \underline{77.36} & 93.58 & 99.11 & 96.25 & \textbf{98.64} & 99.57 & 76.43 & 89.36 & 1.47 & 49.95 \\
    \rowcolor{blue!10}
    EMR-Merging~\cite{huang2024emr}   & \ding{51}          & 71.02 & 72.75 & 93.49 & 99.24 & 96.86 & 98.12 & 99.58 & 74.36 & 88.18 & 2.77  & 15.62 \\ 
     \rowcolor{blue!10}
    FREE-Merging~\cite{zheng2025free}   & \ding{51}          & 74.00 & 77.23 & 93.48 & 99.55 & 96.30 & 98.21 & 99.57 & 75.47 & 89.74 & 1.05  & 10.00 \\ 
    \rowcolor{blue!10}
    TALL-Mask~\cite{wang2024localizing} & \ding{51}  & 73.02 & \textbf{77.38} & 97.63 & 99.38 & 97.15 & 98.46 & 99.66 & 77.61 & 90.04 & 0.71  & 15.62 \\ 
    \rowcolor{blue!10}
    T-Switch~\cite{qi2024less} & \ding{51}  & \underline{74.05} & 77.32 & 96.47 & 99.52 & \underline{97.33} & 98.41 & 99.56 & 78.55 & 90.15 
 & 0.59   & 6.25  \\
    \rowcolor{green!15}
    DTS-T  & \ding{51}  &  \textbf{74.15} & 76.85 & \underline{97.92} & \underline{99.66 }& 97.00 & 98.34 & 99.63 & \textbf{79.03} & \underline{90.32} & \underline{0.40} & 3.68 \\
    \rowcolor{green!15}
    DTS-D & \ding{51}  & \textbf{74.15} & 76.87 & \textbf{97.98} & \textbf{99.78} & \textbf{97.36} & \underline{98.63} & \textbf{99.67} & \underline{78.78} & \textbf{90.40} & \textbf{0.39} & 3.68 \\
    \rowcolor{green!15}
    DTS-T*  & \ding{51}  & 74.01 & 76.13 & 97.74 & 99.52 & 96.96 & 98.15 & 99.60 & 78.51 & 90.08 
 & 0.67  & 0.98  \\
    \rowcolor{green!15}
    DTS-D* & \ding{51}  &  73.97 & 76.42 & 97.58 & 99.56 & 97.13 & 98.58 & 99.47 & 78.63 & 90.17 
  & 0.57 & 0.98 \\ 
    \bottomrule
  \end{tabular}
  }
\end{table*}

\begin{table*}[t]
  \centering
    \caption{Multi-task performance (\%) when merging ViT-L/14 models on eight tasks. ADR refers to the Accuracy Drop Rate, and AMR denotes the Additional Memory Rate. The best result is highlighted in \textbf{bold}, and the second-best result is \underline{underlined}.
  \textbf{Per} indicates whether the merging is a personalized method.}
    \label{table_vit-l-14}
  \resizebox{0.9\textwidth}{!}{
  \begin{tabular}{lccccccccccccc}
    \toprule \rowcolor{gray!10}
    Method & \textbf{Per} & SUN397 & Cars & RESISC45 & EuroSAT & SVHN & GTSRB & MNIST & DTD 
      & Avg. $\uparrow$ & ADR $\downarrow$ & AMR $\downarrow$\\   
    \midrule
    \rowcolor{cyan!10}
    Individual & -      & 81.72 & 92.39 & 98.85 & 99.88 & 98.11 & 99.24 & 99.69 & 84.15 & 94.25 & -- & --      \\
    \midrule
    \rowcolor{purple!10} 
    Weight-Averaging~\cite{wortsman2022model} & \ding{55} &  72.19 & 81.42 & 82.55 & 91.93 & 78.08 & 70.76 & 97.14 & 62.95 & 79.63 & 15.52 & 0.00 \\ 
    \rowcolor{purple!10}
    Fisher-Merging~\cite{matena2022merging} & \ding{55} &  69.24 & 88.61 & 87.50 & 93.53 & 80.66 & 74.82 & 93.32 & 70.07 & 82.22 & 12.77 & 0.00 \\ 
    \rowcolor{purple!10} 
    RegMean~\cite{jin2022dataless} & \ding{55} &  73.38 & 81.80 & 86.10 & 97.01 & 88.12 & 84.27 & 98.54 & 60.82 & 83.76 & 11.14 & 0.00 \\ 
    \rowcolor{purple!10} 
    Task-Arithmetic~\cite{ilharco2022editing} & \ding{55} &  73.92 & 82.13 & 87.64 & 92.82 & 87.91 & 86.77 & 98.94 & 65.64 & 84.47 & 10.38 & 0.00 \\ 
    \rowcolor{purple!10} 
    Ties-Merging~\cite{yadav2023ties} & \ding{55} &  74.74 & 84.50 & 89.00 & 94.18 & 85.66 & 82.07 & 98.65 & 67.71 & 84.56 & 10.28 & 0.00 \\ 
    \rowcolor{purple!10} 
    DARE~\cite{yu2024language} & \ding{55} &  73.03 & 82.70 & 86.19 & 93.41 & 85.26 & 83.48 & 98.58 & 65.69 & 83.54 & 11.36 & 0.00 \\ 
    \rowcolor{purple!10} 
    AdaMerging~\cite{yang2023adamerging} & \ding{55} &  79.03 & 90.34 & 90.86 & 96.19 & 93.44 & 98.05 & 99.12 & 79.94 & 90.87 & 3.59 & 0.00 \\ 
    \rowcolor{purple!10} 
    AdaMerging++~\cite{yang2023adamerging} & \ding{55} &  79.46 & 90.38 & 91.66 & 97.47 & 93.42 & 97.55 & 99.05 & 79.20 & 91.02 & 3.43 & 0.00 \\ \midrule
    \rowcolor{blue!10} 
    SVD  & \ding{51} &  78.49 & 89.73 & 98.02 & 99.86 & 98.08 & 98.91 & 99.71 & 81.91 & 93.09  &1.23 & 4.08 \\   
    \rowcolor{blue!10} 
    Twin-Merging~\cite{lu2024twin} & \ding{51} &  \textbf{ 81.92 } & 91.59 & 96.87 & 99.72 & 98.03 & 92.42 & 99.57 & 83.94 & 93.01 & 1.31 & 100.0 \\ 
    \rowcolor{blue!10} 
    WEMOE~\cite{tang2024merging} & \ding{51}   &  81.42 & 92.10 & 95.46 & 99.48 & 97.73 & 99.13 & 99.70 & 83.74 & 93.60 & 0.70 & 58.80 \\ 
    \rowcolor{blue!10} 
    EMR-Merging~\cite{huang2024emr} & \ding{51} &  80.47 & 90.71 & 98.55 & 99.54 & 97.94 & 99.10 & 99.69 & 82.71 & 93.59 & 0.71 & 15.62 \\ 
     \rowcolor{blue!10}
    FREE-Merging~\cite{zheng2025free} & \ding{51}  & 81.60 & 92.03 & 96.54 & 99.61 & 98.10 & 98.79 & 99.72 & 82.80 &  93.65 &  0.63 & 10.00  \\ 
    \rowcolor{blue!10} 
    TALL-Mask~\cite{wang2024localizing} & \ding{51} &  80.58 &	91.61 &	98.68 &	99.76 &	98.08 &	\underline{99.24} &	\underline{99.74} &	83.14 &	93.85 & 0.42 & 15.62 \\ 
    \rowcolor{blue!10} 
    T-Switch~\cite{qi2024less} & \ding{51} &  \underline{ 81.84 } & \textbf{ 92.38 } & \textbf{ 98.89 } & 99.74 & 98.03 & 99.08 & 99.63 & 83.72 & 94.16 & 0.10 & 6.25 \\
        \rowcolor{green!15} 
    DTS-T & \ding{51} &  81.82 & 91.93 & 98.82 & \textbf{ 99.90 } & 98.05 & 99.21 & \textbf{ 99.77 } & 83.93 & \underline{ 94.18 } & \underline{0.08} & 2.95 \\ 
    \rowcolor{green!15} 
    DTS-D & \ding{51} &  81.75 & 92.09 & \textbf{ 98.89 } & \underline{ 99.88 } & \textbf{ 98.27 } & \textbf{ 99.25 } & { 99.72 } & \textbf{ 84.15 } & \textbf{ 94.24 } & \textbf{0.02} & 2.95 \\ 
    \rowcolor{green!15} 
    DTS-T* & \ding{51} &  81.72 & \underline{ 92.17 } & \underline{ 98.83 } & 99.82 & 98.03 & 99.15 & 99.71 & 83.87 & 94.16 & 0.10 & 0.99 \\ 
    \rowcolor{green!15} 
    DTS-D* & \ding{51} &  81.69 & 91.73 & 98.77 & 99.86 & \underline{ 98.15 } & { 99.22 } & { 99.72 } & \underline{ 83.99 } & 94.14 & 0.12 & 0.99 \\ 
    \bottomrule
  \end{tabular}
  }
\end{table*}

\section{Experiments}
\label{sec: experiment}
In this section, we evaluate both the effectiveness and efficiency of the proposed method through conventional multi-task model merging experiments, and further assess its ability to generalize to unseen tasks.

\subsection{Conventional Multi-task Model Merging}
\label{sec: Conventional}
\textbf{Backbones and datasets}. 
We conduct experiments on visual classification, natural language processing, and natural language generation tasks.
For visual classification, we evaluate our method using three variants of CLIP~\cite{radford2021learning}: ViT-B/32, ViT-L/14, and ViT-B/16. Following prior work~\cite{lu2024twin, qi2024less}, the ViT-B/32 and ViT-L/14 backbones are tested across eight benchmark datasets: SUN397~\cite{xiao2010sun}, Cars~\cite{krause20133d}, RESISC45~\cite{cheng2017remote}, EuroSAT~\cite{helber2019eurosat}, SVHN~\cite{netzer2011reading}, GTSRB~\cite{stallkamp2011german}, MNIST~\cite{deng2012mnist}, and DTD~\cite{cimpoi2014describing}. Additionally, we extend our evaluation to a broader suite of 30 tasks using the ViT-B/16 architecture.
For natural language processing tasks, we adopt RoBERTa~\cite{liu2019roberta} and GPT-2~\cite{radford2019language} as backbones, and evaluate on eight tasks from the GLUE benchmark~\cite{wang2018glue}: CoLA~\cite{warstadt2019neural}, SST-2~\cite{socher2013recursive}, MRPC~\cite{dolan2005automatically}, STS-B~\cite{cer2017semeval}, QQP~\cite{le2021improve}, MNLI~\cite{williams2017broad}, QNLI~\cite{rajpurkar2016squad}, and RTE~\cite{giampiccolo2007third}.
For natural language generation, we use Qwen-14B~\cite{bai2023qwen} as the backbone and evaluate on MMLU~\cite{hendrycksmeasuring}, TruthfulQA~\cite{lin2022truthfulqa}, and BBQ~\cite{parrish2021bbq}, following prior work~\cite{lu2024twin}. Additional details are provided in Sec.~\ref{Evaluation Metrics Across Benchmarks} of the supplementary material.

\textbf{Baselines.}  
We compare our method against both basic and personalized model merging methods. 
Basic methods—including Weight-Averaging~\cite{wortsman2022model}, Fisher-Merging~\cite{matena2022merging}, RegMean~\cite{jin2022dataless}, Task-Arithmetic~\cite{ilharco2022editing}, Ties-Merging~\cite{yadav2023ties}, DARE~\cite{yu2024language}, and AdaMerging~\cite{yang2023adamerging}—produce a single merged model without storing any task-specific information.
In contrast, personalized approaches retain additional task-specific parameters and include simple SVD, Twin-Merging~\cite{lu2024twin}, WEMOE~\cite{tang2024merging}, EMR-Merging~\cite{huang2024emr}, TALL-Mask~\cite{wang2024localizing}, T-Switch~\cite{qi2024less}, and FREE-Merging~\cite{zheng2025free}. 
Unless otherwise noted, we adopt the settings from T-Switch~\cite{qi2024less} for all baselines.
We report results for both \textbf{DTS-T} and \textbf{DTS-D} using a default sparsity coefficient of $r = 0.3$.
To further evaluate efficiency under strict memory constraints, we also provide results for \textbf{DTS-T*} and \textbf{DTS-D*}, where $r$ is adaptively adjusted to ensure that the additional storage overhead remains below 1\% across all backbones.
Additional details on the baselines can be found in Sec.~\ref{Baselines for seen tasks} of the supplementary material.

\textbf{Metrics}.  
We report both absolute performance (accuracy) and relative performance, measured by the \emph{accuracy drop rate} (\textbf{ADR}), defined as the ratio between the accuracy difference of the merged model and its individually fine-tuned counterpart (the upper bound).
A lower drop rate indicates reduced performance degradation.
To assess memory efficiency, we also report the \emph{additional memory rate} (\textbf{AMR}), which quantifies the extra memory required to store task-specific information per task, beyond the storage of the merged model itself. 
Lower values correspond to more memory-efficient approaches.

\begin{table*}[t]
  \centering
   \caption{Multi-task performance (\%) when merging RoBERTa models on eight tasks. ADR refers to the Accuracy Drop Rate, and AMR denotes the Additional Memory Rate. The best result is \textbf{highlighted}, and the second-best result is \underline{underlined}.
   \textbf{Per} indicates whether the merging is a personalized method.}
    \label{table_roberta}
  \resizebox{\textwidth}{!}{
  \begin{tabular}{l
                  cccc  % Single-Sentence Tasks
                  cc   % Similarity & Paraphrase
                  ccc  % Inference
                  cc % Avg Acc, Avg rate
                  c}  % Additional Memory, Rate
    \toprule \rowcolor{gray!10}
    Method  &  \textbf{Per}    & CoLA   & SST2   & MRPC  & STSB   & QQP    & MNLI   & QNLI   & RTE     & Avg. $\uparrow$
    &ADR $\downarrow$
    & AMR $\downarrow$  \\
    \midrule
\rowcolor{cyan!10} 
Individual  & - &  { 60.18 } & { 94.04 } & 89.22 & { 90.63 } & { 91.41 } & { 87.20 } & { 92.71 } & { 79.06 } & { 85.56 } & -- & -- \\ \midrule
\rowcolor{purple!10} 
Weight-Averaging~\cite{wortsman2022model} & \ding{55}  &  13.96 & 64.11 & 69.36 & 31.84 & 75.36 & 42.19 & 58.70 & 55.23 & 51.34 & 40.00 & 0.00 \\ 
\rowcolor{purple!10} 
RegMean~\cite{jin2022dataless} & \ding{55} &  36.67 & 90.60 & 75.74 & 62.68 & 83.55 & 70.02 & 82.35 & 58.48 & 70.01 & 18.17 & 0.00 \\ 
\rowcolor{purple!10} 
Task-Arithmetic~\cite{ilharco2022editing} & \ding{55} &  18.78 & 85.89 & 79.90 & 74.03 & 83.78 & 59.08 & 69.67 & 62.09 & 66.65 & 22.10 & 0.00 \\ 
\rowcolor{purple!10} 
Ties-Merging~\cite{yadav2023ties} & \ding{55} &  20.48 & 84.40 & 81.13 & 58.19 & 85.70 & 64.65 & 74.81 & 42.96 & 64.04 & 25.15 & 0.00 \\
\rowcolor{purple!10} 
DARE~\cite{yu2024language} & \ding{55} &  9.28 & 77.87 & 77.94 & 30.77 & 79.25 & 39.35 & 71.48 & 62.09 & 56.00 & 34.54 & 0.00 \\ \midrule
\rowcolor{blue!10} 
SVD & \ding{51} &  58.31 & \underline{93.92} & 88.48 & \textbf{90.65} & 87.56 & 85.80 & 92.26 & 65.25 & 82.78  & 3.24 & 4.14 \\ 
\rowcolor{blue!10} 
Twin-Merging~\cite{lu2024twin} & \ding{51} &  59.12 & 93.53 & 88.65 & 72.36 & 89.17 & 84.30 & { 92.32 } & 73.89 & 81.67 & 4.55 & 100.0 \\ 
\rowcolor{blue!10} 
EMR-Merging~\cite{huang2024emr} & \ding{51} &  39.96 & 93.35 & 86.27 & 82.73 & 89.72 & 85.45 & 89.57 & 74.37 & 80.18 & 6.29 & 15.62 \\ 
\rowcolor{blue!10} 
FREE-Merging~\cite{zheng2025free} & \ding{51} &  54.50 & 93.69 & 88.46 & 67.04 & 88.03 & 80.60 & 89.90 & 79.06 & 80.16 & 6.31 & 10.00 \\ 
\rowcolor{blue!10} 
TALL-Mask~\cite{wang2024localizing} & \ding{51} &  45.81 & 93.81 & 88.73 & 88.87 & 88.51& 80.29 & \underline{92.37} & 75.09 & 81.69 & 4.52 & 15.62 \\ 
\rowcolor{blue!10} 
T-Switch~\cite{qi2024less} & \ding{51} &  53.12 & \textbf{ 94.04 } & 89.22 & 90.15 & \textbf{ 91.16 } & \underline{ 87.08 } & \textbf{ 92.57 } & \textbf{ 77.26 } & 84.33 & 1.44 & 6.25 \\ 
\rowcolor{green!15} 
DTS-T & \ding{51} &  59.33 & 93.66 & 89.68 & { 90.59 } & 90.55 & 86.50 & 91.80 & \textbf{ 77.26 } & \underline{ 84.93 } & \underline{0.73} & 3.81 \\ 
\rowcolor{green!15} 
DTS-D & \ding{51} &  \underline{ 59.66 } & 93.69 & 89.71 & \underline{ 90.62 } & \underline{ 90.80 } & \textbf{ 87.09 } & 92.11 & 76.17 & \textbf{ 84.98 }  & \textbf{0.67}  & 3.81 \\
\rowcolor{green!15} 
DTS-T* & \ding{51} &  \textbf{ 59.71 } & 93.35 & \underline{ 89.95 } & 90.54 & 88.87 & 85.33 & 91.69 & \underline{ 76.90 } & 84.54 & 1.18 & 0.88 \\ 
\rowcolor{green!15} 
DTS-D* & \ding{51} &  59.26 & 93.81 & \textbf{ 89.96 } & 90.58 & 89.63 & 86.45 & 92.18 & 76.17 & 84.75 & 0.94 & 0.88 \\ 
\bottomrule
  \end{tabular}
  }
\end{table*}

\begin{table*}[t]
  \centering
  \large
   \caption{Multi-task performance (\%) when merging Qwen-14B models on three tasks. AMR denotes the Additional Memory Rate. The best result is \textbf{highlighted}, and the second-best result is highlighted in \underline{underlined}.
   \textbf{Per} indicates whether the merging is a personalized method.}
    \label{table_qwen}
    \renewcommand{\arraystretch}{0.9}
  \resizebox{0.65\textwidth}{!}{
  \begin{tabular}{l
                  cccc  %Tasks
                  c % Avg Acc, Avg rate
                  c}  % Additional Memory, Rate
    \toprule \rowcolor{gray!10}
    Method   &  \textbf{Per}    & MMLU   & TruthfulQA   & BBQ  &  Avg. $\uparrow$			
    & AMR $\downarrow$  \\
    \midrule
    \rowcolor{cyan!10} 
Individual & - &   68.36  &  54.35 & 93.53 & 72.08  & -- \\ \midrule
\rowcolor{purple!10} 
Weight-Averaging~\cite{wortsman2022model} & \ding{55} &  67.11 & 50.02 & 82.32 & 66.48 &  0.00  \\ 
\rowcolor{purple!10} 
DARE~\cite{yu2024language} & \ding{55} &  67.23 & 51.31 & 83.74 & 67.43  & 0.00 \\ 
\rowcolor{purple!10} 
Task-Arithmetic~\cite{ilharco2022editing} & \ding{55} &  66.63 & 53.38 & 78.24 & 66.08  &  0.00 \\ 
\rowcolor{purple!10} 
Ties-Merging~\cite{yadav2023ties} & \ding{55} &  67.28 & 50.02 & 84.10 & 67.13  & 0.00 \\ \midrule
\rowcolor{blue!10} 
SVD & \ding{51} &  67.99 & 52.45 & 91.71 & 70.72  & 5.39 \\ 
\rowcolor{blue!10} 
Twin-Merging~\cite{lu2024twin} & \ding{51} &  68.07 & 52.38 & 90.73 & 70.39  & 100.0 \\ 
\rowcolor{blue!10} 
EMR-Merging~\cite{huang2024emr} & \ding{51} &  67.94 & 52.50 & 91.02 & 70.49  & 15.62 \\ 
\rowcolor{blue!10} 
FREE-Merging~\cite{zheng2025free} & \ding{51} &  68.13 & 53.91 & 92.54 & 71.52  & 10.00 \\ 
\rowcolor{blue!10} 
T-Switch~\cite{qi2024less} & \ding{51} &  68.05 & 53.72 & 92.50 & 71.42  & 6.25 \\ 
\rowcolor{green!15} 
DTS-T & \ding{51} &  \underline{68.30} & \textbf{54.12} & \underline{92.97} & \underline{71.80} &  3.57  \\ 
\rowcolor{green!15} 
DTS-D & \ding{51} &  \textbf{68.32} & \underline{54.11} & \textbf{92.99} & \textbf{71.81} &  3.57\\
\rowcolor{green!15} 
DTS-T* & \ding{51} &  68.20 & 53.99 & 92.90 & 71.70 &  0.92  \\ 
\rowcolor{green!15} 
DTS-D* & \ding{51} &  68.18 & 53.99 & 92.91 & 71.70 &  0.92 \\ 
\bottomrule
  \end{tabular}
  }
\end{table*}

\textbf{Experimental results}.  
Individual models require storing a fully fine-tuned model per task, and we omit their additional memory usage in comparisons.
Table~\ref{table_vit-b-32} and Table~\ref{table_vit-l-14} present detailed comparisons of model performance and additional memory overhead for visual classification tasks.
In addition, we evaluate our method on 30 datasets using ViT-B/16 as the backbone, with detailed results provided in Table~\ref{table_vit-b-16} of the supplementary material.
The following key observations can be made:
(1) Basic model merging methods perform significantly worse than individual models.
(2) Recent personalized merging approaches improve per-task performance by incorporating task-specific parameters. 
However, these methods often incur substantial memory costs.
For instance, WEMOE~\cite{tang2024merging} requires an extra 58.80\% of the model size per task.
In contrast, our method achieves comparable or superior performance with only $\sim1\%$ additional memory per task.
(3) Our method offers flexibility in balancing performance and memory usage through a single tunable sparsity coefficient, allowing it to adapt to varying deployment constraints. 
In comparison, methods such as EMR-Merging~\cite{huang2024emr} rely on fixed storage budgets and lack adaptability, limiting practical applicability in real-world scenarios.

\begin{table*}[t]
  \centering
  \caption{Generalization results (\%) on two unseen tasks for ViT-B/32 models merged from six tasks. The best result is \textbf{highlighted}, and the second-best result is \underline{underlined}.
  \textbf{Per} indicates whether the merging is a personalized method.}
  \label{tab:unseen_vit}
  \resizebox{0.95\textwidth}{!}{
  \begin{tabular}{lccccccccccccc}
    \toprule\rowcolor{gray!10}
    & \multicolumn{7}{c}{Seen Tasks} & \multicolumn{3}{c}{Unseen Tasks} \\
    \cmidrule(lr){3-9} \cmidrule(lr){10-12}\rowcolor{gray!10}
    Method & \textbf{Per}& SUN397 & Cars & EuroSAT & GTSRB & MNIST & DTD & Avg. & RESISC45 & SVHN & Avg. \\     \midrule
    \rowcolor{cyan!10}
    Individual & - & 74.49 & 77.73 & 99.80 & 98.73 & 99.69 & 79.36 & 88.30 & 98.22 & 97.46 & 97.84 \\ \midrule
    \rowcolor{purple!10} 
    Fisher-Merging~\cite{matena2022merging} & \ding{55} &  68.19 & 67.41 & 86.47 & 67.23 & 81.64 & 58.69 & 71.61 & 60.25 & 42.51 & 51.38  \\ 
    \rowcolor{purple!10} 
    RegMean~\cite{jin2022dataless} & \ding{55} &  69.45 & 70.53 & 97.06 & 86.99 & 98.35 & 67.12 & 81.58 & 50.22 & \underline{ 51.50 } & 50.86  \\ 
    \rowcolor{purple!10} 
    Task-Arithmetic~\cite{ilharco2022editing} & \ding{55} &  65.28 & 63.68 & 87.17 & 76.18 & 94.24 & 56.47 & 73.84 & 52.43 & 45.27 & 48.85  \\ 
    \rowcolor{purple!10} 
    Ties-Merging~\cite{yadav2023ties} & \ding{55} &  68.27 & 65.93 & 81.22 & 70.01 & 89.07 & 56.02 & 71.75 & 60.36 & 47.34 & 53.85  \\ 
    \rowcolor{purple!10} 
    DARE~\cite{yu2024language} & \ding{55} &  69.99 & 69.32 & 72.16 & 55.39 & 84.52 & 56.81 & 68.03 & 51.60 & 49.36 & 50.48  \\ 
    \rowcolor{purple!10} 
    AdaMerging~\cite{yang2023adamerging} & \ding{55} &  69.84 & 72.45 & 95.18 & 95.53 & 98.16 & 70.71 & 83.65 & 48.75 & \textbf{ 60.72 } & 54.74  \\  \midrule
    \rowcolor{blue!10} 
    SVD & \ding{51}  & 71.18 & 71.60 & 99.64 & 98.02 & \underline{99.66} & 77.58 & 86.28 & 60.60 & 23.50 & 42.05 \\  
    \rowcolor{blue!10} 
    Twin-Merging~\cite{lu2024twin} & \ding{51} &  71.76 & 69.20 & 73.24 & 93.37 & { 99.64 } & 72.25 & 79.91 & 52.43 & 45.27 & 48.85  \\ 
    \rowcolor{blue!10} 
    WEMOE~\cite{tang2024merging} & \ding{51} &  \textbf{ 74.32 } & \textbf{ 78.16 } & 98.71 & \textbf{ 98.64 } & 99.57 & 75.13 & 87.42 & 47.39 & 51.37 & 49.38  \\ 
    \rowcolor{blue!10} 
    EMR-Merging~\cite{huang2024emr} & \ding{51} &  71.81 & 74.61 & 99.32 & 98.40 & 99.63 & 75.85 & 86.60 & 28.95 & 49.80 & 39.38  \\ 
    \rowcolor{blue!10} 
    TALL-Mask~\cite{wang2024localizing} & \ding{51} &  73.02 & 77.38 & 99.38 & 98.46 &\underline{99.66} & 77.61 & 87.58 & 52.43 & 45.27 & 48.85  \\ 
    \rowcolor{blue!10} 
    T-Switch~\cite{qi2024less} & \ding{51} &  74.05 & \underline{ 77.32 } & 99.52 & 98.41 & 99.56 & 78.55 & 87.90 & 60.60 & 23.50 & 42.05  \\ 
    \rowcolor{green!15} 
    DTS-T & \ding{51} &  \underline{ 74.15 } & 76.85 & 99.66 & 98.34 & 99.63 & \textbf{ 79.03 } & \underline{ 87.94 } & \underline{ 61.35 } & 49.11 & 55.23  \\ 
    \rowcolor{green!15} 
    DTS-D & \ding{51} &  \underline{ 74.15 } & 76.87 & \textbf{ 99.78 } & \underline{ 98.63 } & \textbf{ 99.67 } & 78.78 & \textbf{ 87.98 } & 60.91 & 49.91 & \underline{ 55.41 }  \\ 
    \rowcolor{green!15} 
    DTS-T* & \ding{51} &  74.14 & 76.26 & 99.70 & 98.25 & 99.61 & 78.74 & 87.78 & 61.32 & 48.92 & 55.12  \\ 
    \rowcolor{green!15} 
    DTS-D* & \ding{51} &  74.07 & 76.62 & \underline{ 99.76 } & 98.58 & \textbf{ 99.67 } & \underline{ 78.83 } & 87.92 & \textbf{ 61.42 } & 49.57 & \textbf{ 55.50 }  \\ 
\bottomrule
  \end{tabular}
  }
\end{table*}

\begin{table*}[t]
  \centering
  \caption{Generalization results (\%) on three unseen tasks when merging GPT-2 models on four tasks. The best result is \textbf{highlighted}, and the second-best result is \underline{underlined}.
  \textbf{Per} indicates whether the merging is a personalized method.}
  \label{tab:unseen_GPT-2}
  \resizebox{0.8\textwidth}{!}{
  \begin{tabular}{lcccccccccccc}
    \toprule \rowcolor{gray!10}
     & \multicolumn{5}{c}{Seen Tasks} & \multicolumn{4}{c}{Unseen Tasks} \\ \cmidrule(lr){3-7} \cmidrule(lr){8-11} \rowcolor{gray!10}
    Method & \textbf{Per} & CoLA & MNLI & MRPC & QNLI & Avg. & QQP & RTE & SST-2 & Avg. \\
    \midrule    
   \rowcolor{cyan!10} 
    Individual & - & 76.80 & 81.99 & 80.39 & 88.27 & 81.86 & 89.64 & 65.34 & 91.17 & 82.05 \\
    \midrule \rowcolor{purple!10} 
    Task-Arithmetic~\cite{ilharco2022editing} & \ding{55} & 68.26 & 68.44 & 72.54 & 60.69 & 67.48 & 67.12 & 43.68 & 50.71 & 53.84 \\ \rowcolor{purple!10} 
    Ties-Merging~\cite{yadav2023ties} & \ding{55} & 63.08 & 79.93 & 34.06 & 71.16 & 62.06 & \underline{70.36} & \underline{57.03} & 51.26 & 59.55 \\ \rowcolor{purple!10} 
    DARE~\cite{yu2024language} & \ding{55} & 63.37 & 65.51 & 69.11 & 57.00 & 63.75 & 68.07 & 42.96 & \underline{51.56} & 54.20 \\
    \midrule \rowcolor{blue!10} 
    SVD & \ding{51} & 76.22 & 81.26 & 79.90 & 87.84 & 81.31 & 63.16 & 52.70 & 50.91 & 55.59 \\     
    \rowcolor{blue!10} 
    Twin-Merging~\cite{lu2024twin} & \ding{51} & 76.27 & 80.03 & 79.65 & 87.53 & 80.87 & 69.73 & 55.29 & 51.14 & 58.72 \\ 
    \rowcolor{blue!10} 
    EMR-MERGING~\cite{huang2024emr} & \ding{51} & 73.63 & 81.74 & 80.14 & 87.04 & 80.64 & 69.94 & 54.87 & 51.12 & 58.64 \\ 
    \rowcolor{blue!10} 
    TALL-Mask~\cite{wang2024localizing} & \ding{51} & 74.78 & 78.59 & 78.43 & 88.15 & 79.98 & 67.12 & 43.68 & 50.71 & 53.84 \\ 
    \rowcolor{blue!10} 
    WEMOE~\cite{tang2024merging} & \ding{51} & 72.30 & 80.24 & 77.59 & 83.06 & 78.30 & 69.16 &{56.93} & 51.02 & 58.37 \\ 
    \rowcolor{blue!10} 
    T-Switch~\cite{qi2024less} & \ding{51} & 76.27 & \textbf{81.87} & 79.90 & \underline{88.57} & 81.65 & 63.16 & 52.70 & 50.91 & 55.59 \\ 
    \rowcolor{green!15}
    DTS-T & \ding{51} & \textbf{76.98} & 81.72 & 79.94 & 87.99 & 81.66 & 70.00 & 56.73 & 50.45 & 59.06 \\ 
    \rowcolor{green!15}
    DTS-D & \ding{51} & 76.69 & 81.58 & \underline{80.33} & \textbf{88.64} & \textbf{81.81} & 70.03 & \textbf{57.65} & 51.03 & \underline{59.57} \\
    \rowcolor{green!15}
    DTS-T* & \ding{51} & \underline{76.72} & \underline{81.84} & 79.94 & 88.15 & 81.66 & 70.11 & 56.49 & 51.45 & 59.35 \\ 
    \rowcolor{green!15}
    DTS-D* & \ding{51} & 76.51 & 81.62 & \underline{80.26} & 88.41 & \underline{81.70} & \textbf{71.12} & 56.93 & \textbf{52.08} & \textbf{60.04} \\ 
    \bottomrule
  \end{tabular}
  }
\end{table*}

For natural language processing tasks, as shown in Table~\ref{table_roberta} and Table~\ref{tab:GPT-2_suppl} in the supplementary material, the results on RoBERTa and GPT-2 follow trends similar to those observed in visual classification tasks.
Notably, our method achieves 99.06\% of the individual model's performance on RoBERTa, requiring only 0.88\% additional memory per task, demonstrating a favorable trade-off between efficiency and effectiveness.
For natural language generation tasks, as shown in Table~\ref{table_qwen}, the results align with those observed in the visual classification and natural language processing tasks.
These results confirm that the task-specific information extracted by DTS is both compact and effective, capturing key task characteristics with minimal storage.

\subsection{Generalization on Unseen Tasks}

\textbf{Baselines and datasets.} 
To evaluate the generalization ability of our method, we conduct experiments on unseen tasks using the same set of baselines as in previous sections, with a few differences. 
The details can be found in Sec.~\ref{Baselines for unseen tasks} of the supplementary material.
The dataset setup remains consistent with previous experiments, with minor modifications to assess generalization.
For visual classification tasks, we designate RESISC45~\cite{cheng2017remote} and SVHN~\cite{netzer2011reading} from the original set of eight datasets as unseen tasks, following the protocol in~\cite{tang2024merging}.
This means their fine-tuned models are not accessed during the merging process. 
The remaining six datasets are treated as seen tasks, and their corresponding fine-tuned models are used in merging.
For natural language processing tasks, we consider QQP~\cite{le2021improve}, RTE~\cite{giampiccolo2007third}, and SST-2~\cite{socher2013recursive} as unseen tasks, while the remaining four tasks are treated as seen tasks.
For natural language generation tasks, we treat BBQ as the unseen task and the remaining tasks as seen.

\textbf{Experimental results}.  
The results for the ViT-B/32 and GPT-2 backbones are presented in Table~\ref{tab:unseen_vit} and Table~\ref{tab:unseen_GPT-2}, while those for ViT-L/14 and Qwen-14B backbones can be found in Table~\ref{tab:unseen_vit_l_14} and Table~\ref{tab:unseen_Qwen-14B_suppl} in the supplementary material.
Our method consistently outperforms all baselines on both seen and unseen tasks. 
Notably, while personalized merging methods generally perform better than basic approaches on seen tasks, they do not necessarily guarantee better generalization.
For example, on the ViT-B/32 backbone, all personalized baselines underperform Ties-Merging~\cite{yadav2023ties} on unseen tasks.
This highlights a key limitation: personalized methods tend to optimize for task-specific performance on seen tasks, which may limit their transferability to unseen tasks.
In contrast, our method demonstrates strong performance on both seen and unseen tasks.
Crucially, this gain incurs negligible overhead: 
Our approach achieves consistent performance improvements in a zero-shot and data-free manner.
Besides, the semantic task embeddings are extracted entirely offline, requiring only a single forward pass of the CLIP text encoder prior to deployment. 
By securing significant improvements without additional training costs or online inference latency, DTS demonstrates a superior performance-efficiency trade-off, confirming its practical viability for cold-start multi-task scenarios.

\subsection{Analysis of DTS}
\begin{figure*}[h]
\centering

\begin{minipage}{0.55\textwidth}
\centering
\setlength{\abovecaptionskip}{0cm}
\includegraphics[width=\linewidth]{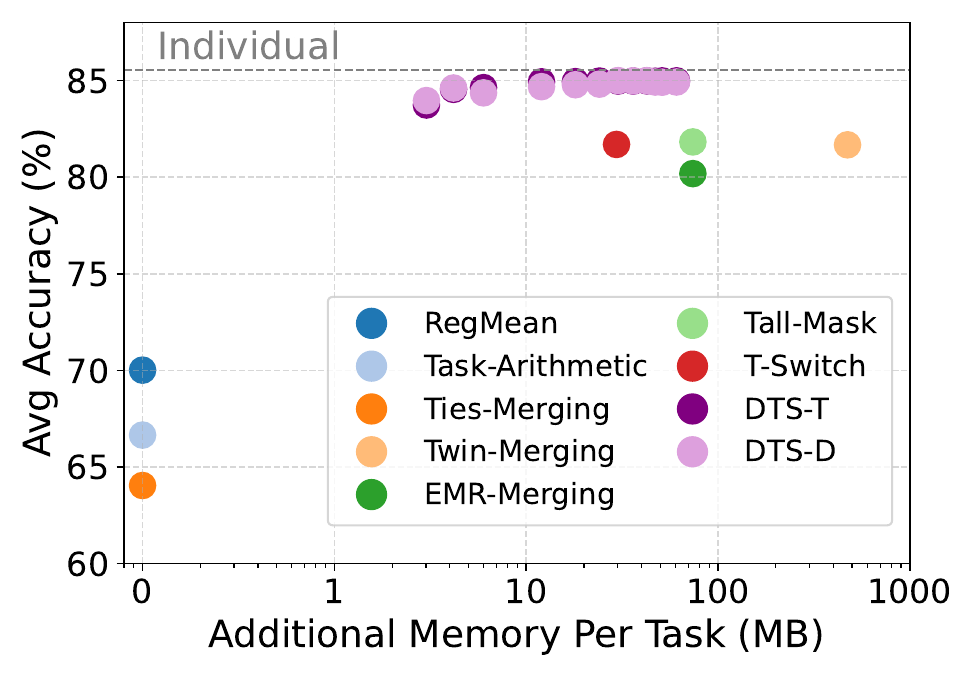}
\caption{Performance (\%) of merged model and additional memory requirement for different methods with RoBERTa as the backbone.}
\label{fig:alpha_tradeoff}
\end{minipage}
\hfill
\begin{minipage}{0.42\textwidth}
\centering
\resizebox{\linewidth}{!}{
\begin{tabular}{lcccc}
\toprule
Method  & QQP & RTE & SST-2 & Avg.\\
\midrule
EMR-Merging~\cite{huang2024emr} & 71.48 & 52.70 & 50.80 & 58.33 \\
WEMOE~\cite{tang2024merging} & \textbf{72.33} & 53.01 & 50.72 & 58.69 \\ 
Twin-Merging~\cite{lu2024twin}  & 71.84 & 56.05 & 50.27 & 59.38 \\
T-Switch~\cite{qi2024less} & \underline{72.15} & 53.42 & 50.91 & 58.83 \\ 
DTS-T  & 70.00 & 56.73 & 50.45 & 59.06 \\
DTS-D & 70.03 & \textbf{57.65} & 51.03 & \underline{59.57}\\
DTS-T* & 70.11 & 56.49 & \underline{51.45} & 59.35 \\ 
DTS-D* & 71.12 & \underline{56.93} & \textbf{52.08} & \textbf{60.04}\\
\bottomrule
\end{tabular}
}
\captionof{table}{Effectiveness of the task-specific information extracted by DTS. Even with the same weighting mechanism, our method consistently achieves the best performance on unseen tasks.}
\label{tab:unseen_auem}
\end{minipage}
\end{figure*}

\textbf{Sensitivity to sparsity factor $r$.}
In our approach, we approximate the parameters of the individual model by retaining only a partial rank of the SVD decomposition, controlled by the sparsity coefficient $r$.
As shown in Fig.~\ref{fig:alpha_tradeoff}, we plot the relationship between performance and model storage overhead for different values of $r$, and compare our method against several baselines.
Additionally, we present the performance of our method for varying values of $r$ in Table~\ref{Ablation r} of the supplementary material.
Our method consistently achieves superior performance with minimal additional storage across all settings.

\textbf{Effectiveness of the task-specific information extracted by DTS.}
To further validate the effectiveness of the task-specific information extracted by our method, we compare it with several existing personalized merging approaches, integrated with our adaptive weighting strategy on unseen tasks.
Specifically, for EMR-Merging~\cite{huang2024emr}, we use the variant of DTS to combine the reconstructed personalized parameters, resulting in a model adapted for unseen tasks. 
For WEMOE~\cite{tang2024merging} and T-Switch~\cite{qi2024less}, we apply our method to the personalized MLP layers and binarized parameters obtained from seen tasks, respectively.
Using the same experimental setup as in the GPT-2 experiments, we report results on unseen tasks in Table~\ref{tab:unseen_auem}. 
Under the same adaptive merging mechanism, our method consistently achieves the best performance, further demonstrating the effectiveness and generalization of the task-specific information.

\begin{table*}[t]
\centering
\begin{minipage}{0.48\textwidth}
\centering
\captionof{table}{The ablation results for Decomposition (D), Thresholding (T), and Scaling (S) when merging ViT-B/32 models.}
\label{ablation}
\resizebox{\linewidth}{!}{
\scriptsize
\begin{tabular}{ccccccccccc}
\toprule
\multirow{2}{*}{D}  &  \multirow{2}{*}{T}  & \multirow{2}{*}{S}  & \multicolumn{2}{c}{DTS-T ($r=0.3$)} & \multicolumn{2}{c}{DTS-T* ($r=0.08$)} \\ 
\cmidrule{4-7} 
& & & Avg. Acc $\uparrow$ & AMR $\downarrow$ & Avg. Acc $\uparrow$ & AMR $\downarrow$ \\ 
\midrule
-- & -- & -- & 90.69 & -- & 90.69 & --  \\
$\checkmark$ & -- & -- & 90.57 & 33.73 & 90.15 & 10.03  \\
-- & $\checkmark$ & $\checkmark$ & 90.41 & 9.37 & 90.41 & 2.52 \\
$\checkmark$ & $\checkmark$ & $\checkmark$ & 90.32 & 3.68 & 90.08 & 0.98 \\
\midrule
\rowcolor{gray!10} $\checkmark$ & $\checkmark$ & -- & 4.98 & 3.69 & 4.80 & 0.99 \\
\bottomrule
\end{tabular}
}

\end{minipage}
\hfill
\begin{minipage}{0.48\textwidth}
\centering

\captionof{table}{The average accuracy (\%) and switching time (ms) of our method against rank-1 LoRA.}
\label{Efficiency}
\renewcommand{\arraystretch}{1.15}
\resizebox{\linewidth}{!}{
\scriptsize

\begin{tabular}{l|c|cc}
\toprule
Backbone & Method & Avg. Acc $\uparrow$ & Time $\downarrow$ \\
\midrule
\multirow{2}{*}{ViT/B-32}
& rank-1 LoRA & 68.61 & 1.01 \\
& Reconstruction & 90.32 & 1.06 \\
\hline
\multirow{2}{*}{ViT/L-14}
& rank-1 LoRA & 81.93 & 1.83 \\
& Reconstruction & 94.18 & 1.97 \\
\hline
\multirow{2}{*}{Qwen-14B}
& rank-1 LoRA & 68.03 & 67.61 \\
& Reconstruction & 71.80 & 69.83 \\
\bottomrule
\end{tabular}
}

\end{minipage}
\end{table*}

\textbf{Effectiveness of each component in DTS. }
Our method is composed of three key components: decomposition, thresholding, and scaling. As shown in Table~\ref{ablation}, without decomposition, storing task-specific information would require significant extra memory.
To demonstrate the effectiveness of thresholding and scaling, we conducted an ablation study, with results presented in Table~\ref{ablation} and Table~\ref{Ablation} of the supplementary material. 
Compared to simple binarization, our thresholding approach yields an accuracy improvement of approximately 2\%. 
Increasing the number of thresholding groups to eight would double the storage cost while providing virtually no additional performance gain.
Additionally, without the scaling strategy, model performance drops to around 5\%. 
These results underscore the effectiveness of each component in our method.

\textbf{Reconstruction Efficiency.}
The reconstruction operation defined in Eq.~ (\ref{reconstruction}) is executed only once upon task switching, strictly conditioned on the task ID. Crucially, the reconstructed weights are cached for all subsequent forward passes. Therefore, DTS introduces absolutely no additional FLOPs or latency during the inference phase.
To evaluate the switching overhead, we compare our reconstruction cost against the standard practice of loading rank-1 LoRA weights. As detailed in Table~\ref{Efficiency}, our method demonstrates a highly favorable trade-off. 
In return, it unlocks substantial performance gains, boosting average accuracy by 21.71\%, 12.25\%,  and 3.77 \% on the respective backbones. This demonstrates that our approach can facilitate robust model merging and dynamic adaptation without compromising deployment efficiency.

\section{Conclusion}
In this work, we revisited the challenge of model merging from the perspective of task similarity and demonstrated that significant performance degradation persists even when merging models fine-tuned on highly similar tasks. 
To address this, we introduced DTS—a compact and effective model merging method based on decomposition, thresholding, and scaling. 
DTS efficiently preserves essential task-specific information by decomposing task vectors into low-rank approximations, achieving high performance with minimal memory overhead.
To support generalization to unseen tasks, we further proposed a variant of DTS, a data-free adaptive merging strategy that weights task-specific information based on the semantic similarity of task characteristics.
Experimental results on standard multi-task model merging benchmarks demonstrate that our method consistently outperforms state-of-the-art baselines, requiring only 1\% extra storage per task. 
Moreover, experiments on unseen tasks show that the adaptive variant of DTS achieves superior generalization performance.

\section*{Acknowledgments}
This work was funded by the National Natural Science Foundation of China under Grants (62276256, U2441251), the Beijing Natural Science Foundation (Z260008), and the National Key Research and Development Program of China (2026ZD1500301).

\bibliographystyle{splncs04}
\bibliography{main}

\clearpage
\setcounter{page}{1}

\section{The Complete Pipeline of DTS}
\label{The Complete Pipeline of DTS}
Here, we take the task vector as an example to illustrate the complete pipeline of DTS, which can also be applied to the difference vector.

We first employ singular value decomposition on
each task vector and retain only the top-r singular values as follows:
\begin{equation}
\boldsymbol{U}_n, \boldsymbol{\Sigma}_n, \boldsymbol{V}_n = \text{SVD}_r(\boldsymbol{\tau}_n),
\end{equation}
where $\boldsymbol{U}_n$ and $\boldsymbol{V}_n$ are the left and right singular vector matrices, and $\boldsymbol{\Sigma}_n$ contains the singular values. Here, r is the proportion of singular values retained.

To further reduce storage, we threshold the decomposed components.
Specifically, take $\boldsymbol{U}_n$ as an example, we mark the positive and negative distribution of all parameters as follows:
\begin{equation}
g(U_{n,j}) = 
\begin{cases}
1, & \text{if } U_{n,j} > 0, \\
0, & \text{otherwise},
\end{cases}
\end{equation}
where $U_{n,j}$ is the $j$-th element of $\boldsymbol{U}_n$.
Then, we divide the positive and negative values into two groups based on their magnitudes as follows:

\begin{equation}
\begin{aligned}
g^{+}(U_{n,j}) &= 
\begin{cases}
\text{1}, & \text{if } U_{n,j} > \lambda_1, \\
\text{0}, & \text{otherwise},
\end{cases} , \\ \quad
g^{-}(U_{n,j}) &= 
\begin{cases}
\text{1}, &  \text{if } \lambda_2 < U_{n,j} < 0,\\
\text{0}, & \text{otherwise},
\end{cases}
\end{aligned}
\end{equation}

where $\lambda_1$ and $\lambda_2$ are the medians of the positive and negative values in $\boldsymbol{U}_n$, respectively.
Finally, a scaling factor is computed for each group:
\begin{equation}
\begin{aligned}
s^+_1 &=  \frac{\left\|\boldsymbol{U}_n \odot g(\boldsymbol{U}_{n}) \odot g^{+}(\boldsymbol{U}_{n})\right\|_2}
              {\left\| g(\boldsymbol{U}_{n}) \odot g^{+}(\boldsymbol{U}_{n})\right\|_2} , \\ \quad
s^+_2 &=  \frac{\left\| \boldsymbol{U}_n \odot g(\boldsymbol{U}_{n}) \odot (1 - g^{+}(\boldsymbol{U}_n))\right\|_2}
              {\left\| g(\boldsymbol{U}_{n}) \odot (1 - g^{+}(\boldsymbol{U}_n))\right\|_2} , \\ \quad
s^-_1 &=  \frac{\left\| \boldsymbol{U}_n \odot g(\boldsymbol{U}_{n}) \odot g^{-}(\boldsymbol{U}_{n})\right\|_2}
              {\left\| g(\boldsymbol{U}_{n}) \odot g^{-}(\boldsymbol{U}_{n})\right\|_2} ,      \\ \quad
s^-_2 &=  \frac{ \left\| \boldsymbol{U}_n \odot g(\boldsymbol{U}_{n}) \odot (1 - g^{-}(\boldsymbol{U}_n))\right\|_2}
              {\left\|g(\boldsymbol{U}_{n}) \odot (1 - g^{-}(\boldsymbol{U}_n))\right\|_2} .              
\end{aligned}
\end{equation}

During inference, the reconstructed $\hat{\boldsymbol{U}}_n$ for each task is computed as follows:
\begin{equation}
\begin{aligned}
\hat{\boldsymbol{U}}_n &= 
s^+_1 \cdot g(\boldsymbol{U}_n) \odot g^{+}(\boldsymbol{U}_n) 
 \\ &+ s^+_2 \cdot g(\boldsymbol{U}_n) \odot (1 - g^{+}(\boldsymbol{U}_n)) 
\\ & + s^-_1 \cdot g(\boldsymbol{U}_n) \odot g^{-}(\boldsymbol{U}_n) 
\\ &+ s^-_2 \cdot g(\boldsymbol{U}_n) \odot (1 - g^{-}(\boldsymbol{U}_n))
\end{aligned}
\end{equation}

Similarly, during the approximation process, we apply the same thresholding strategy to $\boldsymbol{V}_n$, and obtain the approximated $\hat{\boldsymbol{V}}_n$ during inference.
Since $\boldsymbol{\Sigma}_n$ only requires storing its diagonal values, which takes up minimal storage, we do not apply any approximation to $\boldsymbol{\Sigma}_n$.

Finally, during the inference stage, we reconstruct the approximated model parameters for the $n$-th task as follows:
\begin{equation}
\hat{\boldsymbol{\theta}}_n = \boldsymbol{\theta}_0 + \hat{\boldsymbol{U}}_n \times {\boldsymbol{\Sigma}}_n \times \hat{\boldsymbol{V}}_n.
\end{equation}

It is worth noting that in our method, by leveraging singular value decomposition and thresholding-based grouped approximation, we only need to store six mask matrices and eight scaling factors for each task to reconstruct $U_n$ and $V_n$, along with the singular values. 
This significantly reduces the storage cost of task-specific information. 
Moreover, the experimental results demonstrate that our method outperforms all state-of-the-art approaches, needing only 1\% extra storage per task.

\section{Baselines}
\label{Baselines}
\subsection{Baselines for Seen Tasks}
\label{Baselines for seen tasks}
\begin{itemize}[leftmargin=15pt]
\item \textbf{Individual Models} refer to task-specific models before merging.

\item \textbf{Weight-Averaging} performs element-wise averaging of the weights across all fine-tuned models.

\item \textbf{Fisher-Merging}~\cite{matena2022merging} leverages Fisher information matrices to estimate parameter importance and merges model weights using importance-weighted averaging.

\item \textbf{RegMean}~\cite{jin2022dataless} merges models via a closed-form solution to a least-squares problem. 
Given $K$ linear models with weights $\boldsymbol{W}_i$ such that $f_i(\boldsymbol{x}) = \boldsymbol{W}_i^\top \boldsymbol{x}$, the objective is 
\(\min\limits_{\boldsymbol{W}} \sum_{i=1}^{K} \lVert \boldsymbol{W}^\top \boldsymbol{X}_i - \boldsymbol{W}_i^\top \boldsymbol{X}_i \rVert^2\), 
where $\boldsymbol{X}_i$ denotes the input features for the $i$-th model. 
The closed-form solution is:
\begin{equation}
\boldsymbol{W} = \left(\sum_{i=1}^{K} \boldsymbol{X}_i^\top \boldsymbol{X}_i\right)^{-1} \left(\sum_{i=1}^{K} \boldsymbol{X}_i^\top \boldsymbol{X}_i \boldsymbol{W}_i\right).
\end{equation}
The merging process requires computing input inner-product matrices $\boldsymbol{X}_i^\top \boldsymbol{X}_i$ in advance.

\item \textbf{Task-Arithmetic}~\cite{ilharco2022editing} defines task vectors as the difference between fine-tuned model weights and the pre-trained model weights, i.e., $\boldsymbol{\tau}_i = \boldsymbol{\theta}_i - \boldsymbol{\theta}_0$. 
To merge $K$ models $\{\boldsymbol{\theta}_i\}_{i=1}^{K}$, the merged model is computed as:
\begin{equation}
\boldsymbol{\theta}_M = \boldsymbol{\theta}_0 + \lambda \sum_{i=1}^{K} \boldsymbol{\tau}_i,
\end{equation}
where $\lambda$ is a tunable merging coefficient.

\item{\textbf{Ties-Merging}}~\cite{yadav2023ties} attributes performance degradation to conflicts among task vectors and addresses this issue by removing redundant parameters and resolving directional inconsistencies.

\item{\textbf{DARE}}~\cite{yu2024language} leverages the redundancy in language models by randomly dropping a large portion (up to 90\%--99\%) of the delta parameters prior to merging, thereby reducing inter-model interference.

\item{\textbf{AdaMerging}}~\cite{yang2023adamerging} learns merging coefficients in an unsupervised manner, either at the task level (Task-wise AdaMerging) or at the layer level (Layer-wise AdaMerging).
\textbf{AdaMerging++} further incorporates Ties-Merging~\cite{yadav2023ties} as a preprocessing step before learning the coefficients.

\item{\textbf{SVD}} applies singular value decomposition to each task vector matrix and retains only the top-$r$ singular values, while one-dimensional parameter vectors remain unchanged.
To ensure a fair comparison, we control $r$ so that the additional storage required by SVD in the baseline is comparable to that of our method, and we present the performance under these conditions for SVD.

\item{\textbf{Twin-Merging}}~\cite{lu2024twin} stores complete model parameters for each task and employs a routing mechanism that performs weighted parameter fusion based on learned expert distributions.

\item{\textbf{WEMOE}}~\cite{tang2024merging} transforms MLP layers into a mixture-of-experts (MoE) structure via test-time adaptation, while merging the remaining parameters using Task-Arithmetic~\cite{ilharco2022editing}.

\item{\textbf{EMR-Merging}}~\cite{tang2024merging} selects a unified base model through Task-Arithmetic~\cite{ilharco2022editing} and generates lightweight task-specific modulators—including binary masks and scaling factors—to align both the direction and magnitude of each personalized model with the unified base.

\item{\textbf{TALL-Mask}}~\cite{wang2024localizing} uses a data-driven method to identify and eliminate task-irrelevant information from the merged multi-task vector.
The resulting task-specific binary masks are used to localize and extract the knowledge relevant to each individual task.

\item{\textbf{T-Switch}}~\cite{qi2024less} filters out parameters below a predefined threshold, binarizes the remaining ones based on sign, and applies separate scaling factors to positive and negative groups to approximate task-specific contributions.

\end{itemize}

\subsection{Baselines for Unseen Tasks}
\label{Baselines for unseen tasks}
For basic merging methods, the merged model is directly used for inference on both seen and unseen tasks.
For personalized merging methods, we follow the evaluation protocols outlined in the respective original papers for seen tasks. 
For unseen tasks, we adopt the following strategies based on each baseline's design: we use the merged model for Twin-Merging~\cite{lu2024twin} and WEMOE~\cite{tang2024merging}, the unified model for EMR-Merging~\cite{huang2024emr}, and the original pretrained model for simple SVD and T-Switch~\cite{qi2024less}.

\section{More Experimental Details}
\label{Datasets}
\subsection{Storage Constraints and Rank Selection}
The adaptive storage adjustment of our method is a deterministic mechanism that dynamically computes the maximum SVD rank $k$ for every layer $\Delta W \in \mathbb{R}^{d_{in} \times d_{out}}$.
For example, we strictly enforce the 1\% storage limit against the original 32-bit dense weights by solving a precise bit-level constraint as follows:
\vspace{-6pt}
\begin{equation}
\vspace{-6pt}
2k(d_{in} + d_{out}) + 256k \le 0.01 \times 32 \times d_{in} \times d_{out}.
\end{equation}
This yields a direct closed-form solution for the maximum rank and retained rank ratio $r$ as follows:
\vspace{-6pt}
\begin{equation}
\vspace{-4pt}
    k = \left\lfloor \frac{0.16 \times d_{in} \times d_{out}}{d_{in} + d_{out} + 128} \right\rfloor, r = \frac{k}{\min(d_{in}, d_{out})}.
\end{equation}

\subsection{Evaluation Metrics Across Benchmarks}
\label{Evaluation Metrics Across Benchmarks}
For visual classification tasks, we employ classification accuracy as the evaluation metric.
For natural language processing tasks, we follow the standard GLUE evaluation protocol: CoLA is assessed using Matthews correlation, STS-B using the average of Pearson and Spearman correlations, and all other tasks using classification accuracy.
For natural language generation tasks, we use the same evaluation metrics as those in previous work~\cite{lu2024twin}.

\section{More Experimental Results}
\subsection{More Backbones}
In addition to the backbones evaluated in the main paper, we also assess the performance of various methods on ViT-B/16 and GPT-2 backbones.
As shown in Table~\ref{table_vit-b-16} and Tables~\ref{tab:GPT-2_suppl}, our method consistently outperforms the baselines under the conventional multi-task model merging setting, demonstrating the effectiveness and robustness of our approach.

Moreover, for generalization on unseen tasks, we also evaluate the results using ViT-L/14 and Qwen-14B as backbones, as shown in Table~\ref{tab:unseen_vit_l_14} and Table~\ref{tab:unseen_Qwen-14B_suppl}.
Our method consistently outperforms all baselines on both seen and unseen tasks. 

\begin{table}[t]
  \centering
  \caption{Multi-task performance (\%) when merging ViT-B/16 models on 30 tasks. Due to space limitations, we report only the average accuracy across all 30 tasks for each method.
    \textbf{Per} indicates whether the merging is a personalized method.}
  \label{table_vit-b-16}
  \resizebox{0.3\textwidth}{!}{
  \begin{tabular}{lcc}
    \toprule \rowcolor{gray!10}
     Method & \textbf{Per} & Avg. \\ 
    \midrule    
    \rowcolor{cyan!10} 
    Individual & - &  93.05   \\
    \midrule 
    \rowcolor{purple!10} 
    Weight-Averaging~\cite{wortsman2022model} & \ding{55} &  42.54  \\
    \rowcolor{purple!10} 
    Task-Arithmetic~\cite{ilharco2022editing} & \ding{55} & 48.89  \\
    \rowcolor{purple!10} 
    Ties-Merging~\cite{yadav2023ties} & \ding{55} & 37.53 \\ 
    \midrule \rowcolor{blue!10} 
    EMR-MERGING~\cite{huang2024emr} & \ding{51} & 89.54  \\ 
    \rowcolor{blue!10} 
    T-Switch~\cite{qi2024less} & \ding{51}  & 91.96  \\ 
    \rowcolor{green!15}
    DTS-T & \ding{51}  & \underline{92.37}  \\ 
    \rowcolor{green!15}
    DTS-D & \ding{51} & \textbf{92.41} \\
    \rowcolor{green!15}
    DTS-T* & \ding{51} & 91.99  \\ 
    \rowcolor{green!15}
    DTS-D* & \ding{51} & 92.08  \\ 
    \bottomrule
  \end{tabular}
  }
\end{table}

\begin{table*}[!htbp]
  \centering
   \caption{Multi-task performance (\%) when merging GPT-2 models on seven tasks. ADR refers to the Accuracy Drop Rate, and AMR denotes the Additional Memory Rate. The best result is highlighted in \textbf{bold}, and the second-best result is \underline{underlined}.
     \textbf{Per} indicates whether the merging is a personalized method.
     }
   \label{tab:GPT-2_suppl}
  \resizebox{0.8\textwidth}{!}{
  \begin{tabular}{lccccccccccc}
    \toprule \rowcolor{gray!10}
    Method & \textbf{Per}
      & CoLA   & MNLI   & MRPC   & QNLI   & QQP    & RTE    & SST-2  % Single metrics
      & Avg. $\uparrow$
      & ADR $\downarrow$
      & AMR $\downarrow$ \\
    \midrule
    \rowcolor{cyan!10} 
Individual & - &  { 76.80 } & { 81.99 } & 80.39 & 88.27 & { 89.64 } & 65.34 & { 91.17 } & { 81.94 } & -- & -- \\ \midrule
\rowcolor{purple!10} 
Weight-Averaging~\cite{wortsman2022model} & \ding{55} &  55.03 & 55.16 & 51.21 & 57.65 & 76.71 & 44.76 & 52.53 & 56.15 & 31.48 & 0.00 \\ 
\rowcolor{purple!10} 
Fisher-Merging~\cite{matena2022merging} & \ding{55} &  54.81 & 58.12 & 39.53 & 63.28 & 81.46 & 49.12 & 64.74 & 58.72 & 28.34 & 0.00 \\ 
\rowcolor{purple!10} 
RegMean~\cite{jin2022dataless} & \ding{55} &  61.69 & 70.44 & 65.37 & 69.71 & 78.83 & 56.10 & 79.74 & 68.84 & 15.99 & 0.00 \\ 
\rowcolor{purple!10} 
Task-Arithmetic~\cite{ilharco2022editing} & \ding{55} &  68.71 & 68.64 & 69.56 & 70.42 & 81.83 & 47.21 & 83.62 & 70.00 & 14.58 & 0.00 \\ 
\rowcolor{purple!10} 
Ties-Merging~\cite{yadav2023ties} & \ding{55} &  68.48 & 71.46 & 68.49 & 69.56 & 82.57 & 47.68 & 81.82 & 70.01 & 14.56 & 0.00 \\
\rowcolor{purple!10} 
DARE~\cite{yu2024language} & \ding{55} &  67.59 & 65.41 & 72.54 & 62.07 & 79.74 & 44.76 & 72.46 & 66.37 & 19.01 & 0.00 \\ \midrule
\rowcolor{blue!10} 
SVD & \ding{51}  &  74.49 & 80.91 & 79.41 & 88.12 & 88.21 & 64.14 & 91.16 & 80.92  & 1.24 & 3.80
 \\ 
\rowcolor{blue!10} 
Twin-Merging~\cite{lu2024twin}  & \ding{51} &  76.02 & 78.75 & 78.93 & 87.17 & 87.58 & 62.82 & 90.16 & 80.20 & 2.12 & 100.0 \\ 
\rowcolor{blue!10} 
EMR-Merging~\cite{huang2024emr} & \ding{51} &  72.77 & 81.08 & 79.16 & 84.84 & 88.11 & \underline{ 66.43 } & 90.25 & 80.38 & 1.91 & 17.41 \\ 
\rowcolor{blue!10} 
TALL-Mask~\cite{wang2024localizing} & \ding{51} &  74.78 & 78.59 & 78.43 & 88.15 & 85.61 & \textbf{67.50} & 90.82 & 80.55 & 1.82 & 17.41 \\ 
\rowcolor{blue!10} 
T-Switch~\cite{qi2024less} & \ding{51} &  76.27 & \textbf{ 81.87 } & 79.90 & \textbf{ 88.57 } & 88.54 & 63.17 & 90.82 & 81.31 & 0.78 & 6.23 \\ 
\rowcolor{green!15} 
DTS-T & \ding{51} &  \textbf{ 76.98 } & 81.72 & 79.94 & 87.99 & \underline{ 89.20 } & 65.34 & \textbf{ 91.16 } & \underline{ 81.76 } & \underline{0.22} & 5.24 \\ 
\rowcolor{green!15} 
DTS-D & \ding{51} &  \underline{ 76.69 } & 81.58 & \textbf{ 80.63 } & 88.24 & \textbf{ 89.46 } & 64.98 & \underline{ 91.05 } & \textbf{ 81.80 } & \textbf{0.17} & 5.24 \\  
\rowcolor{green!15} 
DTS-T* & \ding{51} &  76.60 & \underline{ 81.86 } & 79.94 & 87.95 & 88.18 & 64.62 & 90.13 & 81.33 & 0.75 & 0.93 \\ 
\rowcolor{green!15} 
DTS-D* & \ding{51} &  76.40 & 81.68 & \underline{ 80.41 } & \underline{ 88.44 } & 88.81 & { 66.06 } & 90.94 & 81.39  & 0.67 & 0.93 \\ 
   \bottomrule
  \end{tabular}
  }
\end{table*}

\begin{table*}[t]
  \centering
  \caption{Generalization results (\%) on two unseen tasks when merging ViT-L/14 models on six tasks. The best result is highlighted in \textbf{bold}, and the second-best result is \underline{underlined}.
    \textbf{Per} indicates whether the merging is a personalized method.}
  \label{tab:unseen_vit_l_14}
  \resizebox{0.9\textwidth}{!}{
  \begin{tabular}{lccccccccccccc}
    \toprule \rowcolor{gray!10}
     & & \multicolumn{7}{c}{Seen Tasks} & \multicolumn{3}{c}{Unseen Tasks}  \\ \cmidrule(lr){3-9} \cmidrule(lr){10-12}\rowcolor{gray!10}
  Method & \textbf{Per}   & SUN397 & Cars & EuroSAT & GTSRB & MNIST & DTD & Avg & RESISC45 & SVHN & Avg. \\
    
    \midrule
    \rowcolor{cyan!10}
    Individual & -      & 81.72 & 92.39 & 99.88 & 99.24 & 99.69 & 84.15 & 92.85 & 98.85 & 98.11 & 98.48 \\
    \midrule
    \rowcolor{purple!10} 
    Fisher-Merging~\cite{matena2022merging} & \ding{55}  &  68.92 & 88.61 & 94.93 & 82.67 & 90.05 & 72.41 & 82.93 & 70.55 & 64.75 & 67.65  \\ 
    \rowcolor{purple!10} 
    RegMean~\cite{jin2022dataless} & \ding{55} &  77.83 & 89.92 & 98.53 & 92.62 & 98.90 & 78.71 & 89.42 & 59.94 & 75.52 & 67.73  \\ 
    \rowcolor{purple!10} 
    Task-Arithmetic~\cite{ilharco2022editing} & \ding{55} &  75.28 & 85.70 & 95.50 & 89.37 & 98.91 & 69.20 & 85.66 & 69.02 & 69.89 & 69.46  \\ 
    \rowcolor{purple!10} 
    Ties-Merging~\cite{yadav2023ties} & \ding{55} &  77.21 & 88.37 & 96.74 & 92.44 & 99.33 & 74.63 & 88.12 & 66.00 & 72.32 & 69.16  \\ 
    \rowcolor{purple!10} 
    DARE~\cite{yu2024language} & \ding{55} &  76.56 & 84.45 & 97.00 & 90.46 & 99.58 & 71.12 & 86.53 & 71.87 & 65.80 & 68.84  \\ 
    \rowcolor{purple!10} 
    AdaMerging~\cite{yang2023adamerging} & \ding{55} &   79.98 & 90.34 & 97.14 & 98.47 & 99.30 & 80.52 & 90.96 & 64.18 & \textbf{ 78.57 } & 71.38  \\ \midrule
    \rowcolor{blue!10} 
    SVD & \ding{51}  & 78.49 & 89.73 & 99.86 & 98.91 &  99.71 & 81.91  & 91.43&  71.33 & 58.45 & 64.89 \\
    \rowcolor{blue!10} 
    Twin-Merging~\cite{lu2024twin} & \ding{51} &  \textbf{ 81.95 } & 91.60 & 99.73 & 93.42 & 99.59 & 83.54 & 91.64 & 69.02 & 69.89 & 69.46  \\ 
    \rowcolor{blue!10} 
    WEMOE~\cite{tang2024merging} & \ding{51} &  81.64 & 92.22 & 99.53 & 99.14 & 99.68 & 82.85 & 92.51 & 61.36 & \underline{ 76.58 } & 68.97  \\ 
    \rowcolor{blue!10} 
    EMR-Merging~\cite{huang2024emr} & \ding{51} &  80.87 & 91.52 & 99.64 & 99.10 & { 99.73 } & 83.24 & 92.35 & 60.03 & 70.60 & 65.32  \\ 
    \rowcolor{blue!10} 
    TALL-Mask~\cite{wang2024localizing} & \ding{51}  &  80.58 & 91.61 & 99.76 & \underline{99.24} &\underline{99.74} & 83.14 & 92.34 & 69.02 & 69.89 & 69.46  \\ 
    \rowcolor{blue!10} 
    T-Switch~\cite{qi2024less} & \ding{51} &  \underline{ 81.84 } & \textbf{ 92.38 } & 99.74 & 99.08 & 99.63 & 83.72 & 92.73 & 71.33 & 58.45 & 64.89  \\ 
    \rowcolor{green!15} 
    DTS-T & \ding{51} &  81.82 & 91.93 & \textbf{ 99.90 } & 99.21 & \textbf{ 99.77 } & 83.93 & 92.76 & \underline{ 72.90 } & 71.80 & 72.35  \\ 
    \rowcolor{green!15} 
    DTS-D & \ding{51} &  81.75 & 92.09 & \underline{ 99.88 } & \textbf{ 99.25 } & 99.72 & \textbf{ 84.15 } & \textbf{ 92.81 } & 72.58 & 72.01 & 72.30  \\
    \rowcolor{green!15} 
    DTS-T* & \ding{51} &  81.72 & \underline{ 92.24 } & 99.78 & 99.12 & 99.67 & 83.85 & 92.73 & \textbf{ 72.98 } & 71.95 & \underline{ 72.47 }  \\ 
    \rowcolor{green!15} 
    DTS-D* & \ding{51} &  81.79 & 92.07 & 99.86 & { 99.22 } & 99.72 & \underline{ 83.99 } & \underline{ 92.78 } & 72.75 & 72.69 & \textbf{ 72.72 }  \\ 
    \bottomrule
  \end{tabular}
  }
\end{table*}

\begin{table}[t]
  \centering
  \caption{Generalization results (\%) on unseen tasks BBQ when merging Qwen-14B models on two seen tasks. The best result is highlighted in \textbf{bold}, and the second-best result is \underline{underlined}.
    \textbf{Per} indicates whether the merging is a personalized method.}
  \label{tab:unseen_Qwen-14B_suppl}
  \resizebox{0.7\textwidth}{!}{
  \begin{tabular}{lccccc}
    \toprule \rowcolor{gray!10} & 
     & \multicolumn{3}{c}{Seen Tasks} & Unseen Task \\ \cmidrule(lr){3-5} \cmidrule(lr){6-6} \rowcolor{gray!10}
    Method   & \textbf{Per} & MMLU & TruthfulQA & Avg. & BBQ  \\
    \midrule    
    \rowcolor{cyan!10} 
    Individual  & -   & 68.36 & 54.35 & 61.36 & 94.53  \\
    \midrule 
    \rowcolor{purple!10} 
    DARE~\cite{yu2024language} & \ding{55}  & 67.82 & 52.79 & 60.31 & 85.96 \\
    \rowcolor{purple!10} 
    Task-Arithmetic~\cite{ilharco2022editing} & \ding{55}  & 67.05 & 53.52 & 60.29 & 84.88  \\
    \rowcolor{purple!10} 
    Ties-Merging~\cite{yadav2023ties} & \ding{55}  & 67.74 & 51.46 & 59.60 & 85.01 \\ 
    \midrule \rowcolor{blue!10} 
    Twin-Merging~\cite{lu2024twin} & \ding{51}  & 68.14 & 52.78 & 60.46 & 86.46  \\ 
    \rowcolor{blue!10} 
    EMR-MERGING~\cite{huang2024emr} & \ding{51} & 68.00 & 52.91 & 60.46 & 86.11  \\ 
    \rowcolor{blue!10} 
    T-Switch~\cite{qi2024less} & \ding{51} & 68.05 & 53.72 & 60.89 & 80.69  \\ 
    \rowcolor{green!15}
    DTS-T  & \ding{51}  & \underline{68.30} & \textbf{54.12} & \underline{61.21} & 87.12 \\ 
    \rowcolor{green!15}
    DTS-D & \ding{51}  & \textbf{68.32} & \underline{54.11} & \textbf{61.22} & 87.09 \\
    \rowcolor{green!15}
    DTS-T* & \ding{51} & 68.20 & 53.99 & 61.10 & \underline{87.28} \\ 
    \rowcolor{green!15}
    DTS-D* & \ding{51} & 68.10 & 53.99  & 61.05 & \textbf{87.30}  \\ 
    \bottomrule
  \end{tabular}
  }
\end{table}

\subsection{Dynamic Thresholding}
We implemented a dynamic thresholding approach using K-Means thresholding on the ViT-B/32 backbone as shown in Table~\ref{Dynamic_Thresholding}.
The results demonstrate that although K-Means thresholding can better capture task-specific features on seen tasks, it does not yield universal improvements on unseen tasks. 
\begin{table}[!htbp]
  \centering
 \caption{Comparison of different thresholding approaches on the ViT-B/32 backbone.}
 \label{Dynamic_Thresholding}
  \resizebox{0.7\textwidth}{!}{
  \begin{tabular}{cccc}
    \toprule \rowcolor{gray!10}
    Method  & \textbf{Thresholding} & Avg. Acc (Seen Tasks) & Avg. Acc (Unseen Tasks)  \\
    \midrule 
    \multirow{2}{*}{DTS-T} & Median & {87.94}  & 55.23   \\
                 & K-Means & \textcolor{red!95!black}{88.25} & \textcolor{red!95!black}{55.76}  \\
    \midrule 
    \multirow{2}{*}{DTS-D} & Median & 87.98  & 55.41   \\ 
                 & K-Means  & \textcolor{red!95!black}{88.13}  & \textcolor{green!70!black}{55.37} \\
    \bottomrule
  \end{tabular}
  }
\end{table}

\subsection{Ablation on Decomposition, Thresholding, and Scaling Strategies}
Our method primarily consists of decomposition, thresholding, and scaling strategies. 
Without decomposition, as shown in Table~\ref{Ablation_decomposition}, storing task-specific information would require
substantial additional memory, although it would not lead to a performance drop. 
In the thresholding strategy, we first apply sign-based thresholding to the model parameters, followed by separate thresholding of the positive and negative components, each further divided into two groups.
This design is intended to preserve finer-grained task-specific information.
To evaluate the effectiveness of this thresholding strategy, we conduct an ablation study comparing our proposed method with a baseline of binarization using ViT-B/32 as the backbone.
The binarization strategy divides model parameters into positive and negative components, assigning a single scaling factor to each.
As shown in Table~\ref{Ablation}, our fine-grained thresholding significantly outperforms the single-step approach, demonstrating its advantage in retaining valuable task-specific information.
Additionally, to demonstrate the effectiveness of our scaling strategy, we conduct an ablation study, as shown in Table~\ref{Ablation}. Without the scaling strategy, the model performance drops significantly, highlighting the importance of our approach.

\begin{table*}[!htbp]
    \centering
   \caption{Ablation results (\%) on decomposition strategies with ViT-B/32 as the backbone. AMR denotes the Additional Memory Rate.}
   \label{Ablation_decomposition}
   \resizebox{\textwidth}{!}{
    \begin{tabular}{lcccccccccc}
    \toprule \rowcolor{gray!10}
    Method  & SUN397 & Cars & RESISC45 & EuroSAT & SVHN & GTSRB & MNIST & DTD    & Avg. $\uparrow$   & AMR $\downarrow$ \\ \midrule   \rowcolor{blue!10} 
    \rowcolor{green!15}
    DTS-T &  74.15 & 76.85 & 97.92 & 99.66 & 97.00 & 98.34 & 99.63 & 79.03 & 90.32 & 3.68 \\
    \rowcolor{blue!10} 
    Without Decomposition & 74.39  &76.51	& 98.15	 &99.80	 &	97.33  & 98.51 & 99.61 &	78.94  & 90.41 & 9.37 \\
    \rowcolor{green!15}
    DTS-D & 74.15 & 76.87 & 97.98 & 99.78 & 97.36 & 98.63 & 99.67 & 78.78 & 90.40  & 3.68 \\
    \rowcolor{blue!10} 
    Without Decomposition  & 74.11 & 76.82 &	98.16 &	99.78 &	97.44 &	98.57 &	99.67 & 78.94 & 90.43  & 9.37	 \\

      \bottomrule
    \end{tabular}
    }
\end{table*}

\begin{table*}[!htbp]
    \centering
   \caption{Ablation results (\%) on thresholding and scaling strategies with ViT-B/32 as the backbone. Without thresholding, we employ binarization as an alternative. }
   \label{Ablation}
   \resizebox{0.9\textwidth}{!}{
    \begin{tabular}{lccccccccc}
    \toprule \rowcolor{gray!10}
    Method  & SUN397 & Cars & RESISC45 & EuroSAT & SVHN & GTSRB & MNIST & DTD & Avg.   \\ \midrule   \rowcolor{blue!10} 
    \rowcolor{green!15}
    DTS-T &  74.15 & 76.85 & 97.92 & 99.66 & 97.00 & 98.34 & 99.63 & 79.03 & 90.32  \\
    \rowcolor{blue!10} 
    Threshold into 8 groups & 74.15 & 76.88 & 97.93 & 99.64 & 97.00 & 98.33 & 99.63 & 79.00 & 90.32  \\
    \rowcolor{blue!10} 
    Without Thresholding & 71.85 &	71.98 &	97.21 &	99.50 &	95.77 &	97.83 &	99.48 &	74.51 &	88.52 \\
    \rowcolor{purple!10} 
    Without Scaling &  0.43	&  0.58	&  3.19	&  10.64	&  6.95	&  3.50	&  10.92	&  3.65	& 4.98 \\
    \rowcolor{green!15}
    DTS-D & 74.15 & 76.87 & 97.98 & 99.78 & 97.36 & 98.63 & 99.67 & 78.78 & 90.40  \\
    \rowcolor{blue!10} 
    Threshold into 8 groups & 74.15 & 76.88 & 97.98 & 99.78 & 97.38 & 98.63 & 99.67 & 78.90 &  90.42 \\
    \rowcolor{blue!10} 
    Without Thresholding  & 72.53 &	72.00 &	97.56 &	99.68 &	95.98 &	97.95 &	99.59 &	74.71 &	88.75 \\
    \rowcolor{purple!10} 
    Without Scaling  & 0.44 &	0.81 &	3.88 &	11.65 &	10.71 &	5.68 &	10.70 &	3.05 &	5.87 \\ 
      \bottomrule
    \end{tabular}
    }
\end{table*}

\begin{table*}[!htbp]
    \centering
   \caption{Sensitivity Analysis (\%) on the sparse coefficient $r$. The table reports the average accuracy on the benchmark datasets for the ViT-B/32 and RoBERTa backbones, respectively.}
   \label{Ablation r}
   \resizebox{.85\textwidth}{!}{
    \begin{tabular}{lcccccccccccc}
    \toprule \rowcolor{gray!10}
    $r$ & 0.05 & 0.07 & 0.1 & 0.2 & 0.3 & 0.4 & 0.5 & 0.6 & 0.7 & 0.8  & 0.9 & 1.0   \\ \midrule  
     \rowcolor{cyan!10}
    \rowcolor{green!15} 
      ViT/B-32 & 89.68 & 90.08 & 90.18 & 90.25 & 90.32 & 90.37 & 90.41 & 90.40 & 90.39 & 90.40 & 90.40 & 90.41    \\    \rowcolor{green!15} 
       RoBERTa & 83.73 & 84.49 & 84.58 & 84.63 & 85.93 & 84.94 & 84.97 & 84.97 & 84.98 & 84.96 & 90.40 & 84.98 \\  
      \bottomrule
    \end{tabular}
    }
\end{table*}

\subsection{Sensitivity Analysis on Sparse Factor r}
As shown in Table~\ref{Ablation r}, we report the accuracy of DTS-D under varying sparsity factors $r$ on the ViT-B/32 and RoBERTa backbones.
Notably, when $r < 0.5$, increasing the number of preserved parameters consistently improves performance.
However, for $r > 0.5$, the gains begin to plateau, indicating diminishing returns with additional parameter retention.
These results suggest that, for practical deployments, setting $r$ within the range $[0, 0.5]$ is sufficient to achieve strong performance while maintaining low memory overhead.
Finally, in practical deployments, setting $r$ within the range $[0, 0.5]$ is sufficient to achieve strong performance while maintaining low memory overhead.

\subsection{Merging Models with the Same Tasks}
To further support the observation made in the main paper—that even for similar tasks, merged models often exhibit substantial performance gaps compared to individually fine-tuned models—we provide additional analysis below.
As shown in Table~\ref{tab: same task 1_suppl}, the performance of the merged model is significantly lower than that of the fine-tuned model on both digit classification and single-sentence tasks.
This underscores the importance of preserving task-specific information during the model merging process.

\begin{table*}[!htbp]
  \centering
  \caption{Merging results (\%) on different datasets with the same task under various backbones.}
  \label{tab: same task 1_suppl}
  \resizebox{\textwidth}{!}{
  \begin{tabular}{l
                  ccc
                  ccc
                  cccc   % Similarity & Paraphrase
                  }
    \toprule \rowcolor{gray!10}
    & \multicolumn{3}{c}{Digit Classification (ViT/B-32)}
    & \multicolumn{3}{c}{Digit Classification (ViT/L-14)}
    & \multicolumn{4}{c}{Similarity \& Paraphrase Tasks (RoBERTa)}\\ 
    \cmidrule(lr){2-4}
    \cmidrule(lr){5-7}
    \cmidrule(lr){8-11} \rowcolor{gray!10}
    Method
      & MNIST   & SVHN   & Avg.  
      & MNIST   & SVHN   & Avg.  
      & MRPC   & STSB  & QQP & Avg.  \\
    \midrule  \rowcolor{cyan!10} 
    Individual   & 99.69 & 97.46 & 98.58 & 99.69 & 98.11 & 98.90   & 89.22 & 90.63	& 91.41 & 90.42
    \\ \rowcolor{purple!10} 
    DARE~\cite{yu2024language}    & 96.15 & 88.68 & 92.42 &  99.48 & 94.62 & 97.05 & 81.57 & 55.21 & 81.94 & 72.91
     \\ \rowcolor{purple!10} 
    Task-Arithmetic~\cite{ilharco2022editing}    & 99.41 & 92.85 & 96.13 & 95.80 & 97.66 & 96.13 & 83.82 & 68.08 & 85.53 & 79.14
     \\ \rowcolor{purple!10} 
    Ties-Merging~\cite{yadav2023ties}   & 99.29 & 93.14 & 96.22 & 99.45 & 94.51 & 96.98 &  82.84 & 61.75 & 87.49 & 77.36
  \\ \rowcolor{green!15} 
      DTS-T   & 99.70 & 97.44 & 98.57 & 99.77 & 98.06 & 98.91 &  89.96 & 90.60 & 91.14 & 90.56
  \\\rowcolor{green!15} 
      DTS-D   & 99.69 & 97.46 & 98.57 & 99.73 & 99.27 & 99.50 &  89.72 & 90.63 & 90.91 & 90.42
  \\
    \bottomrule
  \end{tabular}
  }

\end{table*}

\begin{table*}[!htbp]
  \centering
  \caption{Comparison of performance (\%) between task vector (-task) and difference vector (-diff) applications to baseline methods for merging ViT-B/32 models on eight tasks. ({\color{red}{The values}}) represent the performance gains.}
  \label{table_vit-b-32_diff_suppl}
  \resizebox{\textwidth}{!}{
  \begin{tabular}{lccccccccc}
    \toprule \rowcolor{gray!10}
    Method & SUN397 & Cars & RESISC45 & EuroSAT & SVHN & GTSRB & MNIST & DTD & Avg. \\
    \midrule
    \rowcolor{cyan!10}
    Individual    & 74.49 & 77.73 & 98.22 & 99.80 & 97.46 & 98.73 & 99.69 & 79.36 & 90.69   \\ 
    \rowcolor{cyan!10}
    Weight Averaging~\cite{wortsman2022model}  & 65.35 & 63.41 & 71.42 & 71.69 & 64.20 & 52.82 & 87.56 & 50.18 & 65.83     \\ \midrule
    \rowcolor{purple!10}
    Task-Arithmetic~\cite{ilharco2022editing}-task &  54.78 & 54.98 & 67.69 & 78.70 & 80.21 & 69.68 & 97.34 & 50.37 & 69.22 \\ 
    \rowcolor{green!20}
    Task-Arithmetic~\cite{ilharco2022editing}-diff & 64.65 & 63.26 & 72.10 & 71.88 & 64.16 & 52.80 & 87.46 & 50.74 & 65.88 ({\color{red}{-3.34}})      \\
    \rowcolor{purple!10}
    Ties-Merging~\cite{yadav2023ties}-task        & 64.17 & 64.43 & 76.31 & 76.62 & 81.28 & 69.37 & 96.53 & 54.52 & 72.90   \\
    \rowcolor{green!20}
    Ties-Merging~\cite{yadav2023ties}-diff         & 65.65 & 63.56 & 71.82 & 68.44 & 62.83 & 51.62 & 87.90 & 51.60 & 65.43 ({\color{red}{-7.47}})      \\
    \rowcolor{purple!10}
    EMR-Merging~\cite{huang2024emr}-task             & 71.02 & 72.75 & 93.49 & 99.24 & 96.86 & 98.12 & 99.58 & 74.36 & 88.18 \\
    \rowcolor{green!20}
    EMR-Merging~\cite{huang2024emr}-diff             & 74.73 & 77.98 & 98.00 & 99.72 & 97.27 & 98.73 & 99.66 & 79.15 & 90.66 ({\color{red}{+2.48}}) \\ 
    \rowcolor{purple!10}
    T-Switch~\cite{qi2024less}-task &  {74.05} & {77.32} & 96.47 & 99.52 & {97.33} & 98.41 & 99.56 & 78.55 & 90.15  \\
    \rowcolor{green!20}
    T-Switch~\cite{qi2024less}-diff   & 74.72 & 77.85 & 98.13 & 99.70 & 97.39 & 98.85 & 99.69 & 79.68 & 90.75 ({\color{red}{+0.60}}) \\  
    \rowcolor{purple!10}
    DTS-task&  {74.15} & {76.85} & {97.92} & {99.66}& 97.00 & 98.34 & {99.63} & {79.03} & {90.32} \\
    \rowcolor{green!20}
    DTS-diff & {74.15} & {76.87} & {97.98} & {99.78} & {97.36} & {98.63} & {99.67} & {78.78} & {90.40} ({\color{red}{+0.08}}) \\
    \bottomrule
  \end{tabular}
  }
\end{table*}

\begin{table*}[!htbp]
  \centering
    \caption{Comparison of performance (\%) between task vector (-task) and difference vector (-diff) applications to baseline methods for merging ViT-L/14 models on eight tasks. ({\color{red}{The values}}) represent the performance gains.}
    \label{table_vit-l-14_diff_suppl}
  \resizebox{\textwidth}{!}{
  \begin{tabular}{lcccccccccc}
    \toprule \rowcolor{gray!10}
    Method & SUN397 & Cars & RESISC45 & EuroSAT & SVHN & GTSRB & MNIST & DTD 
      & Avg. \\ \midrule
    \rowcolor{cyan!10}
    Individual      & 81.72 & 92.39 & 98.85 & 99.88 & 98.11 & 99.24 & 99.69 & 84.15 & 94.25  \\     
      \rowcolor{cyan!10}
    Weight Averaging~\cite{wortsman2022model} &  72.19 & 81.42 & 82.55 & 91.93 & 78.08 & 70.76 & 97.14 & 62.95 & 79.63  \\  \midrule
     \rowcolor{purple!10} 
    Task-Arithmetic~\cite{ilharco2022editing}-task &  73.92 & 82.13 & 87.64 & 92.82 & 87.91 & 86.77 & 98.94 & 65.64 & 84.47  \\ 
     \rowcolor{green!20}
    Task-Arithmetic~\cite{ilharco2022editing}-diff &  72.39 & 82.54 & 82.83 & 92.42 & 78.23 & 70.65 & 97.01 & 62.55 & 79.83 ({\color{red}{-4.64}})  \\ 
    \rowcolor{purple!10} 
    Ties-Merging~\cite{yadav2023ties}-task &  74.74 & 84.50 & 89.00 & 94.18 & 85.66 & 82.07 & 98.65 & 67.71 & 84.56  \\ 
    \rowcolor{green!20} 
    Ties-Merging~\cite{yadav2023ties}-diff &  73.11 & 81.92 & 83.25 & 90.28 & 77.31 & 66.41 & 96.82 & 62.93 & 79.00 ({\color{red}{-5.56}}) \\ 
    \rowcolor{purple!10} 
    EMR-Merging~\cite{huang2024emr}-task &  80.47 & 90.71 & 98.55 & 99.54 & 97.94 & 99.10 & 99.69 & 82.71 & 93.59  \\ 
    \rowcolor{green!20} 
    EMR-Merging~\cite{huang2024emr}-diff &  81.78 & 92.31 & 98.85 & 99.90 & 98.12 & 99.23 & 99.75 & 83.67 & 94.20 ({\color{red}{+0.61}}) \\ 
    \rowcolor{purple!10} 
    T-Switch~\cite{qi2024less}-task &  81.84 & 92.38 & 98.89 & 99.74 & 98.03 & 99.08 & 99.63 & 83.72 & 94.16  \\ 
    \rowcolor{green!20} 
    T-Switch~\cite{qi2024less}-diff &  81.95 & 92.69 & 98.92 & 99.9 & 98.13 & 99.23 & 99.73 & 84.57 & 94.39 ({\color{red}{+0.23}}) \\ 
    \rowcolor{purple!10} 
    DTS-task &  {81.82} & 91.93 & 98.82 & { 99.90 } & 98.05 & 99.21 & { 99.77 } & 83.93 & { 94.18 }  \\ 
    \rowcolor{green!20} 
    DTS-diff &  {81.75} & {92.09} & { 98.89 } & { 99.88 } & { 99.27 } & { 99.25 } & { 99.72 } & { 84.15 } & { 94.24 } ({\color{red}{+0.06}})  \\ 
    \bottomrule
  \end{tabular}
  }
\end{table*}

\begin{table*}[!htbp]
  \centering
   \caption{Comparison of performance (\%) between task vector (-task) and difference vector (-diff) applications to baseline methods for merging RoBERTa models on eight tasks. ({\color{red}{The values}}) represent the performance gains.}
   \label{tab:roberta-2_diff_suppl}
  \resizebox{\textwidth}{!}{
  \begin{tabular}{l
                  ccc  % Single-Sentence Tasks
                  cc   % Similarity & Paraphrase
                  ccc  % Inference
                  c % Avg Acc, Avg rate
                }  % Additional Memory, Rate
    \toprule \rowcolor{gray!10}
    Method      & CoLA   & SST2   & MRPC  & STSB   & QQP    & MNLI   & QNLI   & RTE     & Avg.  \\
    \midrule
    \rowcolor{cyan!10} 
Individual &  { 60.18 } & { 94.04 } & 89.22 & { 90.63 } & { 91.41 } & { 87.20 } & { 92.71 } & { 79.06 } & { 85.56 }  \\ 
\rowcolor{cyan!10} 
Weight Averaging~\cite{wortsman2022model} &  13.96 & 64.11 & 69.36 & 31.84 & 75.36 & 42.19 & 58.70 & 55.23 & 51.34  \\  \midrule
\rowcolor{purple!10} 
Task-Arithmetic~\cite{ilharco2022editing}-task &  18.78 & 85.89 & 79.90 & 74.03 & 83.78 & 59.08 & 69.67 & 62.09 & 66.65  \\ 
\rowcolor{green!20} 
Task-Arithmetic~\cite{ilharco2022editing}-diff &  9.28 & 79.47 & 77.70 & 31.80 & 79.41 & 40.03 & 72.36 & 61.01 & 56.38 ({\color{red}{-10.27}}) \\ 
\rowcolor{purple!10} 
Ties-Merging~\cite{yadav2023ties}-task &  20.48 & 84.40 & 81.13 & 58.19 & 85.70 & 64.65 & 74.81 & 42.96 & 64.04  \\ 
\rowcolor{green!20} 
Ties-Merging~\cite{yadav2023ties}-diff &  33.16 & 80.28 & 73.53 & 10.85 & 81.28 & 49.09 & 65.60 & 57.04 & 56.35 ({\color{red}{-7.69}}) \\ 
\rowcolor{purple!10} 
EMR-Merging~\cite{huang2024emr}-task &  39.96 & 93.35 & 86.27 & 82.73 & 89.72 & 85.45 & 89.57 & 74.37 & 80.18  \\ 
\rowcolor{green!20} 
EMR-Merging~\cite{huang2024emr}-diff &  51.70 & 93.46 & 89.22 & 88.17 & 91.19 & 87.13 & 92.24 & 75.45 & 83.57 ({\color{red}{+3.39}}) \\ 
\rowcolor{purple!10}
T-Switch~\cite{qi2024less}-task &  53.12 & 94.04  & 89.22 & 90.15 &  91.16  &  87.08  &  92.57  &  77.26  & 84.33  \\ 
\rowcolor{green!20} 
T-Switch~\cite{qi2024less}-diff &  50.76 & 93.81 & 89.22 & 89.35 & 91.17 & 87.34 & 92.62 & 76.17 & 83.80 ({\color{red}{-0.53}})  \\ 
\rowcolor{purple!10}
DTS-task &  59.33 & 93.66 & 89.68 & { 90.59 } & 90.55 & 86.50 & 91.80 & { 77.26 } & { 84.93 }  \\ 
\rowcolor{green!20} 
DTS-diff &  { 59.66 } & 93.69 & 89.71 & { 90.62 } & { 90.80 } & { 87.09 } & 92.11 & 76.17 & { 84.98 }  ({\color{red}{+0.05}}) \\
    \bottomrule
  \end{tabular}
  }
\end{table*}

\begin{table*}[!htbp]
  \centering
   \caption{Comparison of performance (\%) between task vector (-task) and difference vector (-diff) applications to baseline methods for merging GPT-2 models on eight tasks. ({\color{red}{The values}}) represent the performance gains.}
   \label{tab:GPT-2_diff_suppl}
  \resizebox{\textwidth}{!}{
  \begin{tabular}{lcccccccc}
    \toprule \rowcolor{gray!10}
    Method
      & CoLA   & MNLI   & MRPC   & QNLI   & QQP    & RTE    & SST-2  % Single metrics
      & Avg. \\
    \midrule
    \rowcolor{cyan!10} 
Individual &  { 76.80 } & { 81.99 } & 80.39 & 88.27 & { 89.64 } & 65.34 & { 91.17 } & { 81.94 }  \\ 
\rowcolor{cyan!10} 
Weight-Averaging~\cite{wortsman2022model} &  55.03 & 55.16 & 51.21 & 57.65 & 76.71 & 44.76 & 52.53 & 56.15  \\  \midrule
\rowcolor{purple!10} 
Task-Arithmetic~\cite{ilharco2022editing}-task &  68.71 & 68.64 & 69.56 & 70.42 & 81.83 & 47.21 & 83.62 & 70.00  \\ 
\rowcolor{green!20} 
Task-Arithmetic~\cite{ilharco2022editing}-diff &  55.03 & 59.24 & 50.98 & 57.60 & 76.69 & 44.76 & 52.52 & 56.69 ({\color{red}{-13.31}}) \\ 
\rowcolor{purple!10} 
Ties-Merging~\cite{yadav2023ties}-task &  68.48 & 71.46 & 68.49 & 69.56 & 82.57 & 47.68 & 81.82 & 70.01 \\ 
\rowcolor{green!20} 
Ties-Merging~\cite{yadav2023ties}-diff &  59.82 & 66.04 & 54.16 & 53.77 & 80.23 & 48.01 & 50.91 & 58.99 ({\color{red}{-11.02}}) \\ 
\rowcolor{purple!10} 
EMR-Merging~\cite{huang2024emr}-task &  72.77 & 81.08 & 79.16 & 84.84 & 88.11 &  66.43  & 90.25 & 80.38  \\ 
\rowcolor{green!20} 
EMR-Merging~\cite{huang2024emr}-diff &  74.01 & 81.80 & 80.14 & 88.33 & 89.47 & 67.14 & 90.82 & 81.67 ({\color{red}{+1.29}})  \\ 
\rowcolor{purple!10} 
T-Switch~\cite{qi2024less}-task &  76.27 &  81.87  & 79.90 &  88.57  & 88.54 & 63.17 & 90.82 & 81.31 \\ 
\rowcolor{green!20} 
T-Switch~\cite{qi2024less}-diff &  75.83 & 82.19 & 80.88 & 88.79 & 89.59 & 67.50 & 91.16 & 82.28  ({\color{red}{+0.97}}) \\
\rowcolor{purple!10} 
DTS-task  &  { 76.98 } & 81.72 & 79.94 & 87.99 & { 89.20 } & 65.34 & { 91.16 } & { 81.76 }  \\ 
\rowcolor{green!15} 
DTS-diff &  { 76.69 } & 81.58 & { 80.63 } & 88.24 & { 89.46 } & 64.98 & { 91.05 } & { 81.80 }  ({\color{red}{+0.04}}) \\  
    \bottomrule
  \end{tabular}
  }
\end{table*}

\subsection{Extending Difference Vector to Scenarios Where the Pre-trained Model Is Inaccessible}
Most prior model merging methods rely on task vectors, which require access to both pre-trained model parameters and fine-tuned model parameters. 
However, in scenarios where the pre-trained model is inaccessible—a more realistic and common setting—these methods become inapplicable.
In contrast, this paper introduces the \emph{difference vector}, which naturally extends to such cases.
Specifically, we first obtain the merged model $\boldsymbol{\theta}_m$ via simple weight averaging over $N$ fine-tuned models $\{\boldsymbol{\theta}_n\}_{n=1}^{N}$, and then compute the difference vector as $\boldsymbol{d}_n = \boldsymbol{\theta}_n - \boldsymbol{\theta}_m$.
When the pre-trained model is inaccessible, the difference vector serves as a practical substitute for the task vector used in previous methods.

As shown in Tables~\ref{table_vit-b-32_diff_suppl},~\ref{table_vit-l-14_diff_suppl},~\ref{tab:roberta-2_diff_suppl}, and~\ref{tab:GPT-2_diff_suppl}, applying the difference vector to existing methods consistently yields performance that significantly surpasses simple weight averaging.
Moreover, we compare difference vectors with task vectors and find that for conventional model merging methods, difference vectors result in slightly lower performance than task vectors, but still outperform naive averaging.
In contrast, personalized model merging methods benefit from using difference vectors, often achieving higher accuracy than when using task vectors.
Notably, for certain backbones, personalized models merged using difference vectors even surpass the performance of individually fine-tuned models. 
These results suggest that, compared to task vectors, the proposed difference vector provides a more accurate and transferable representation of task-specific information, particularly in the absence of a pre-trained model.

\end{document}